\documentclass[journal,table]{IEEEtran}
\usepackage{afterpage}
\usepackage{amsmath}
\usepackage{changes}
\usepackage{graphicx}
\usepackage{color}
\usepackage[numbers,sort&compress]{natbib}
\usepackage{url}
\usepackage{xcolor}
\usepackage{collcell}
\usepackage{hhline}
\usepackage{pgf}
\usepackage{booktabs}
\usepackage{tabularx}
\usepackage{gincltex}
\usepackage{balance}
\usepackage{multirow}
\usepackage{subfig}
\usepackage{pgfplots}
\usepackage{pgfkeys}
\usepackage{tikzscale}
\usepackage{siunitx}
\usepackage{tikz}
\usepackage[utf8]{inputenc}
\pgfplotsset{compat=newest}
\usepackage{booktabs}
\usetikzlibrary{pgfplots.groupplots}
\definecolor{myred}{rgb}{.8,.0,.0}

\DeclareMathOperator*{\fun}{func} 

\def\colorModel{hsb}
\newcommand\ColCell[1]{
  \pgfmathparse{#1<110?1:0}  
    \ifnum\pgfmathresult=0\relax\color{white}\fi
  \pgfmathsetmacro\compA{220/360}      
  \pgfmathsetmacro\compB{(#1/200)} 
  \pgfmathsetmacro\compC{1}      
  \edef\x{\noexpand\centering\noexpand\cellcolor[\colorModel]{\compA,\compB,\compC}}\x #1
  } 
\newcolumntype{E}{>{\collectcell\ColCell}m{1.6 cm}<{\endcollectcell}}

\newcommand\ColCellH[1]{
  \pgfmathparse{#1<520?1:0}  
    \ifnum\pgfmathresult=0\relax\color{white}\fi
  \pgfmathsetmacro\compA{220/360}      
  \pgfmathsetmacro\compB{(#1/1000)} 
  \pgfmathsetmacro\compC{1}      
  \edef\x{\noexpand\centering\noexpand\cellcolor[\colorModel]{\compA,\compB,\compC}}\x #1
  } 
\newcolumntype{H}{>{\collectcell\ColCellH}m{1.6 cm}<{\endcollectcell}}

\pgfkeys{
  /includetikzgraphic/.is family, /includetikzgraphic,
  default/.style = {
    width = \textwidth,
    ratio = 0.8},
  width/.estore in = \itgWidth,
  ratio/.estore in = \itgRatio
}

\newlength\figureheight
\newlength\figurewidth

\begin{document}

\bstctlcite{IEEEexample:BSTcontrol}
\title{Primary Tumor Origin Classification of Lung Nodules in Spectral CT using Transfer Learning}


\author{
    \IEEEauthorblockN{L.S.~Hesse\IEEEauthorrefmark{1,2}, P.A. de Jong\IEEEauthorrefmark{3}, J.P.W. Pluim\IEEEauthorrefmark{2,3}, V. Cheplygina\IEEEauthorrefmark{2}}\\
    \IEEEauthorblockA{\IEEEauthorrefmark{1}University of Oxford, United Kingdom}\\
    \IEEEauthorblockA{\IEEEauthorrefmark{2}Eindhoven University of Technology, The Netherlands}\\
    \IEEEauthorblockA{\IEEEauthorrefmark{3}Utrecht Medical Center, The Netherlands}\\
}

\maketitle

\begin{abstract} 
Early detection of lung cancer has been proven to decrease mortality significantly. A recent development in computed tomography (CT), spectral CT, can potentially improve diagnostic accuracy, as it yields more information per scan than regular CT.  However, the shear workload involved with analyzing a large number of scans drives the need for automated diagnosis methods. Therefore, we propose a detection and classification system for lung nodules in CT scans. Furthermore, we want to observe whether spectral images can increase classifier performance. For the detection of nodules we trained a VGG-like 3D convolutional neural net (CNN). To obtain a primary tumor classifier for our dataset we pre-trained a 3D CNN with similar architecture on nodule malignancies of a large publicly available dataset, the LIDC-IDRI dataset. Subsequently we used this pre-trained network as feature extractor for the nodules in our dataset. The resulting feature vectors were classified into two (benign/malignant) and three (benign/primary lung cancer/metastases) classes using support vector machine (SVM). This classification was performed both on nodule- and scan-level. We obtained state-of-the art performance for detection and malignancy regression on the LIDC-IDRI database. Classification performance on our own dataset was higher for scan- than for nodule-level predictions. For the three-class scan-level classification we obtained an accuracy of 78\%. Spectral features did increase classifier performance, but not significantly. Our work suggests that a pre-trained feature extractor can be used as primary tumor origin classifier for lung nodules, eliminating the need for elaborate fine-tuning of a new network and large datasets. Code is available at \url{https://github.com/tueimage/lung-nodule-msc-2018}.
\end{abstract}

\begin{IEEEkeywords}
Spectral computed tomography, lung nodules, computer-aided detection and diagnosis, convolutional neural network, transfer learning 

\end{IEEEkeywords}

 \section{Introduction}

\IEEEPARstart{L}{ung} cancer is the leading cause of death among all cancer patients for both men and women~\citep{americancancersociety}. Five year survival ratings for not metastasized cancer vary between 13\% and 92\% depending on the stage of the cancer when diagnosed~\citep{americancancersociety}. Therefore, early and accurate diagnosis is crucial in increasing the patients' prospect of survival. The National Lung Screening Trial (2011) showed that screening patients with low dose computed tomography (CT) decreases mortality from lung cancer~\citep{NationalScreening}. The obtained CT images must be analyzed by a radiologist, who detects the presence of lung nodules in order to interpret the scan. Lung nodules are round or oval shape growths in the lungs which can be either malignant, indicating lung cancer, or benign, such as a calcification or inflammation.

A new development in CT acquisition, spectral CT, could increase the available information obtained in one scan, potentially resulting in more accurate patient diagnosis. In spectral CT (also called dual-energy CT) two CT scans are acquired simultaneously with different energy spectra~\citep{Johnson2012}. The spectral CT used in this study is a detector based spectral CT, in which a dual-layer detector absorbs high and low energy photons separately~\citep{Rassouli2017}. This means patient dose does not have to be increased. From the two energy scans multiple reconstructions can be made to visually extract specific spectral information such as iodine only, non-contrast and effective atomic number images~\citep{Rassouli2017}. Studies on the clinical use of spectral CT suggest that the virtual non-contrast reconstruction could facilitate the assessment of lung nodules as it provides additional information about the degree of contract enhancement and the presence of calcifications~\citep{Kang2010,chae2008clinical}. However, the resulting number of images can be large and combining all this information into a correct diagnosis might be a challenging task. Therefore, in this study we propose to use the spectral information to develop a computer aided diagnosis system which can classify lung nodules based on the origin of the primary tumor. Early knowledge about the tumor origin from the CT scan could facilitate an immediate start of the most suitable therapy.

Computer aided detection and diagnosis of lung nodules has been a rapidly developing field. Most research has been aimed at the detection and malignancy estimation of nodules~\citep{LUNA2016,anode2009,DeWit2017,Znet,Liu2016,igrt,Hussein2017,Chen2017}. Since the introduction of neural networks in the field, the performance of these systems has improved significantly and is now competing with radiologist’s performance~\cite{LUNA2016}. On the other hand, classification on other factors than malignancy is still a relatively unexplored topic of study. Few studies have differentiated nodules on their appearance (solid, ground glass and partly solid)~\citep{Jacobs2015,Tu2017} and a recent study proposed the classification as benign, primary lung cancer or metastatic lung cancer~\cite{Nishio2018}. To the best of our knowledge, the use of spectral information has not yet been applied in the automated classification of lung nodules. 

For this study we obtained a dataset containing spectral thorax CT scans, for which both patient diagnosis and nodule annotations are known. Annotations are the actual nodule locations marked by one or more radiologists. However, we want to develop a system which can predict patient diagnosis from a scan  without annotations, as these annotations are usually not available. Hence, it is necessary to develop a nodule detector which is able to detect the suspicious areas in the spectral CT scans as a first step. However, the available dataset is relatively small, making it infeasible to train a detection network with this data. Therefore we propose to train a nodule detector on a large publicly available database containing thorax CT scans with nodule annotations (\mbox{LIDC-IDRI \citep{Armato2015}}), and apply this detector to the spectral CT scans.

Next, we propose a classification method which can categorize both nodules and scans  by primary tumor origin. We first train a regression network on the LIDC-IDRI database, predicting a malignancy score per nodule, and use this network off-the-shelf as feature extractor for our own database. This process of transferring a classifier from one domain to another can be called \textit{transfer learning}~\cite{cheplygina2018cats}. The obtained features from our database are then classified based on tumor origin using a support vector machine (SVM)~\citep{sharif2014cnn}. The spectral images are added as extra features to observe whether the performance increases from this information. We expect that by adding these images a higher performance can be achieved than by using only the conventional data, as they can contain additional information about the tumor~\citep{Kang2010,chae2008clinical}.  

This paper is organized as follows. Section~\ref{relatedwork} describes the related work. In Section~\ref{datasets} the datasets used in this study are presented. Section~\ref{methods} introduces the proposed methods and section~\ref{experiments} explains how we applied our methods to the datasets in more detail. In section~\ref{results} and~\ref{discussion} the experimental results are presented and discussed. Finally, section~\ref{conclusion} concludes this paper.

\section{Related Work} \label{relatedwork}

\subsection{Definitions}
In this study detection is defined as the localization of lung nodules. Classification is the categorization of a nodule or scan into one of the defined classes. Detection sensitivity is defined as the proportion of all actual nodules which is detected. A false positive (FP) is a predicted nodule location which has no overlap with an actual nodule.  

\subsection{Neural network architectures}
Convolutional neural networks (CNN) are widely used in medical image analysis~\citep{litjens2017survey}. Most currently used neural networks are derived from a few well known architectures developed in the last couple of years. In 2012, the top performance in the ImageNet challenge was significantly improved by a CNN called AlexNet~\citep{krizhevsky2012imagenet}, after which the development of other CNN architectures took off. In subsequent years, winning networks were mostly deeper with smaller receptive fields, such as VGG19, GoogleNet and Resnet~\citep{simonyan2014very,szegedy2015going,he2016deep}. VGG19 consists of 19 convolutional layers followed by three fully connected layers and is frequently used as starting point for classification networks due to its simplicity~\citep{litjens2017survey}.

Segmentation and thus inherently detection, can be either achieved using a classification network with a sliding window, or by using a fully convolutional network. In 2015, U-net was proposed, which is a fully convolutional network consisting of an up- and downsampling part, with `skip' connections to connect low and high scales directly with each other~\citep{ronneberger2015u}. Since its introduction U-net has been widely used for medical image segmentation and detection. 
\renewcommand{\arraystretch}{1.2}
\begin{table*}\centering
\caption{Characteristics of the spectral dataset from the UMCU. The values between brackets are the standard deviations.  }
\begin{tabular}{@{}lllllll@{}} \toprule

Variables & Total & Benign Multinodular & Benign & Primary Lung & Melanoma & Colorectal   \\ \midrule
Nodules   & 2088  & 868                  & 123   & 264 & 329      & 504     \\
Patients  & 196   & 76                   & 54   & 31  & 15       & 20    \\
CT Scans     & 214   & 78                   & 56  & 32   & 20       & 28   \\

\hspace{3mm}Contrast-Enhanced & 172 &  57 &39  & 29& 20 & 27 \\
 \addlinespace

Nodules per patient & 9.8 (11.7) &  11.1 (10.8) &2.2 (2.66) & 8.3 (10.9) & 16.5 (15.4) & 18.0 (13.8)  \\
Nodule size [mm] & 7.6 (6.9) &  5.8 (4.9) &4.8 (4.8) & 11.5 (6.9)& 8.0 (4.9) & 9.0 (5.4)  \\
\bottomrule
\end{tabular}
\label{table:nodules}
\end{table*}

\subsection{Application to lung nodules }
In the last couple of years lung nodules have received quite some attention due to recent grand challenges concerning lung nodules. In 2016 the LUng Nodule Analysis challenge (LUNA2016) was organized~\citep{setio2017validation}, in which participants had to develop an automated method to detect lung nodules. The availability of a large public dataset of 1018 thorax CT scans containing annotated nodules, the Lung Image Database and Image Database Resource Initiative (LIDC-IDRI), made the organization of this challenge possible. As the same dataset was used, and evaluation for all participants was equal, the challenge provided a thorough analysis of state of the art nodule detection algorithms. It was observed that compared to a similar challenge in 2009 (ANODE2019~\citep{anode2009}), where participants used only classical detection methods, there was a complete shift towards the use of neural networks in the proposed algorithms. The sensitivity was assessed as an average of the sensitivity at 7 false positive rates between 1/8 and 8 FPs per scan. For comparison, it was shown that radiologists achieve a 0.51-0.83 sensitivity, with a false positive rate ranging between 0.33-1.39 per scan~\citep{armato2009assessment}. Most contestants used 2D convolutional networks and a two stage method consisting of suspicious location detection as a first step, and using false positive reduction on these locations to obtain the final predictions. The top performing team in LUNA2016 obtained an average sensitivity of 0.811 using a ResNet based network with increased layer width and decreased depth~\citep{Znet}. This performance was significantly higher than the best score of the previous ANODE2009 challenge (0.632), indicating that neural networks perform very well in the field of nodule detection. Because submission to the challenge was kept open after paper publication, it was shown that an even higher detection sensitivity can be achieved with more advanced neural networks. These networks use mostly 3D filters and predict nodules in a one step approach~\citep{igrt}. 

The achievements in lung nodule detection were followed by an interest in malignancy estimation of the nodules. In 2017, Kaggle organized the National Science Bowl with the main subject being lung cancer predictions. For each CT scan, participants were asked to predict whether the patient was diagnosed with lung cancer within a year. Because patient diagnoses were too distant from the actual cancer features in the image, most participants used nodule detection as a first step in their methods~\citep{igrt,DeWit2017}. The challenge was won by~\citet{igrt}. In their algorithm they used a 3D version of a modified U-net for the detection of suspicious nodules, followed by cancer estimation from the five most suspicious proposals. The final cancer probability was obtained using the feature maps of the last convolutional layer of the network for each of the five proposals, and combining these into one probability~\citep{igrt}. Another approach for estimating the malignancy in this challenge was proposed by~\citet{DeWit2017}. In this system a patch-based convolution 3D network was used, predicting both nodule probability and malignancy simultaneously.
	
Using nodules annotated with a malignancy score~\citet{Liu2016} trained a nodule-level CNN to extract the radiologists knowledge, and transfered these weights to predict patient-level malignancy with multiple instance learning (MIL). Another study  characterized nodule malignancy using nodule shape and image features extracted with a CNN~\cite{buty2016characterization}. In both~\cite{Hussein2017} and~\cite{Chen2017} not only nodule malignancy information was used to predict malignancy scores, but also seven other nodule attributes in a multi-task learning approach. 

Lung nodule classification on other characteristics than malignancy has been performed as well. \citet{Jacobs2015} and \citet{Tu2017} attempted to classify nodules as solid, part-solid or non-solid, reaching a performance similar to radiologists agreement. In~\citep{sharif2014cnn} it was shown that using an off-the-shelf CNN, which is a CNN trained on another dataset, in combination with SVM performed surprisingly well on a number of image classification tasks. In another study this approach was applied to identify nodules as perifissural nodules (PFN) or non PFNs, using the same off-the-shelf network known as OverFeat~\citep{Sermanet2013,ciompi2015automatic}. Very recently \citet{Nishio2018} developed a classification system to differentiate nodules as benign, primary lung cancer or metastatic lung cancer. In that study a 2D CNN was used which was pre-trained on ImageNet and fine-tuned on their dataset.  

\section{Datasets} \label{datasets}
\subsection{LIDC-IDRI}
For this study we made use of the Lung Image Database Consortium and Image Database Resource Initiative (LIDC-IDRI)~\citep{Armato2015}. This database consists of thorax CT images for 1018 cases. Lung nodule annotations are available for each subject in the dataset. Every scan was annotated by four radiologists in a two-phase annotation process. In the first phase each radiologist independently annotated all nodules, classifying each nodule into one of three categories: \mbox{nodule $<$ 3 mm}, \mbox{nodule $\geq$ 3 mm}, \mbox{non-nodule $\geq$ 3 mm}. In the second phase each radiologist was shown their own annotations, along with the annotations of the other reviewers. From these marks they had to come up with their final opinion. For nodules larger than 3 mm, radiologists were instructed to segment the nodules.
Next to the detection of the nodules, the radiologists had to score each nodule on its malignancy (a score between 1 and 5), and seven other characteristics.

In the LUNA2016 challenge it was proposed to use only images with a slice thickness larger than 3 mm, and exclude images with inconsistent slice spacing or missing slices~\citep{setio2017validation}. Furthermore, only \mbox{nodules $\geq$ 3 mm} annotated by at least three out of four radiologists were considered as relevant nodules in the challenge. Non-nodules and \mbox{nodules $<$ 3 mm} were considered irrelevant findings. During evaluation findings overlapping with irrelevant findings were not considered as true positives, nor as false positives. Furthermore, the LUNA2016 challenge constructed a list of nodule candidates using existing classical nodule detection algorithms. In this study, the reduced set of 888 scans from the LUNA2016 challenge is used. Likewise, the relevant nodules and irrelevant findings are adopted as proposed in the challenge.

\subsection{Spectral CT}
The spectral CT images are obtained from the University Medical Center Utrecht (UMCU). Patients were selected by automatically screening diagnosis reports from the spectral CT on the words `lung nodules'. These reports were assessed and categorized on diagnosis by our radiologist with more than 10 years experience. Based on the size of the groups it was chosen to include patients from the following 5 categories: benign, benign multinodular, primary lung cancer, melanoma, and colorectal cancer . This resulted in a dataset of 279 thorax CT scans. Each of the included scans was annotated by our radiologist, who annotated all nodules with a maximum of forty nodules per scan. During annotation the radiologist marked some scans which he considered unsuitable for this study, these scans were excluded from the dataset. An overview of the excluded scans can be found in Appendix \ref{appendix:exclusion}. To be consistent, nodules smaller than 3 mm were excluded as well, as these were too small to be a nodule according to the definition used in the LUNA2016 challenge. Growths larger than 3 cm are referred to as masses according to this definition, however these are kept in the dataset as it is expected that especially large growths contain important information for classification. The final dataset contained 214 scans, more information about the dataset is shown in Table~\ref{table:nodules}.

The spectral data consists for each scan of a conventional multi-energetic view, which is comparable to a regular CT, and of multiple virtual mono-energetic views. With these mono-energetic views it is possible to calculate other spectral reconstructions.

\begin{figure*}
\centering
\includegraphics[width=1\textwidth]{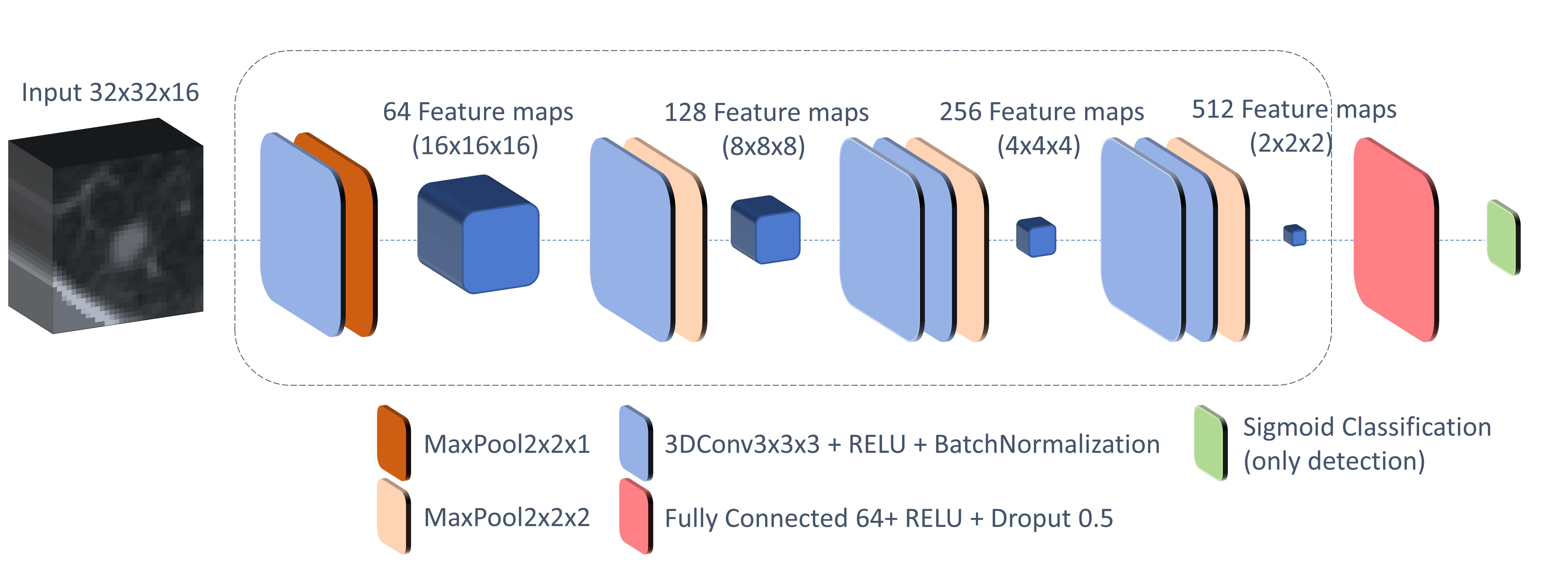}
\caption{Architecture of detection and classification network for a scale of 3.2 cm. The input size is 32x32x16 voxels, corresponding to 3.2 cm as the resolution of the slice thickness is 2 mm. The final sigmoid classification layer is only used for detection, which outputs a probability for each input cube. During classification the network predicts a regression value. For the larger scale of 6.4 cm (detection only), the architecture is identical, with all dimensions of the individual feature maps multiplied by two. The box around part of the network indicates the part which is used as feature extractor for classification.}
\label{fig:architecture}
\end{figure*}

\section{Methods} \label{methods}
Our proposed methodology consists of two main parts, the detection of nodules in the CT scans and subsequently the classification of these nodules, based on either their malignancy or on the origin of the primary tumor. The final aim of performing both detection and classification is to use the detected suspicious locations in the images as input for the classification in an end-to-end pipeline. However, in this study nodules annotated by an experienced radiologist are used as input for the classification as these are available and can provide an independent analysis of the classification method. 

Nodules are 3D round or oval growths in the lungs. As CT images contain 3D information about the nodules shape, we want to preserve this knowledge during computation. For this reason 3D convolutional networks are used in this study for both detection and classification.

The nodule detection algorithm should detect suspicious places in a CT scan, which can then be used as input for a classification algorithm. It is not necessary to obtain an exact nodule location or segmentation, a patch containing the suspicious area is sufficient. For this reason the proposed detection network is a patch-based classification network which labels an image patch as containing a nodule or not. The network architecture is depicted in Figure~\ref{fig:architecture} and will be explained in more detail at the end of this section. During training both positive and negative 3D patches are cropped from the image and given as input to the network. Because nodule detection is a highly unbalanced problem, data augmentation is applied to the positive samples. During evaluation a detection map for the whole image is obtained by analyzing the image in a sliding window fashion. The outcome of every predicted patch is assigned to the middle pixels of the patch, a cube of 8 mm. The size of this cube was chosen to balance spatial resolution and computation time per scan. After each prediction the patch is translated in such way that for each image location probability predictions are generated. The network is trained and evaluated on two different scales individually (3.2 cm and 6.4 cm), as it is expected that the combination of predictions is superior to the individual predictions~\citep{ju2018relative}. The final probabilities for the whole image are obtained by a simple average of the posterior probabilities for each scale. 

The nodule classification aims at labeling nodules based on the origin of the primary tumor in the spectral dataset. However, because some of the patient groups in this dataset are relatively small for classification (see Table~\ref{table:nodules}) a malignancy regression model is first trained on the LIDC-IDRI dataset and then applied to the spectral dataset. It is expected that for predicting a malignancy score and for predicting the location of the primary tumor, similar features could be of importance. For the regression model a comparable patch-based CNN is used as for detection. This model is trained using only image patches containing nodules with varying malignancy, and predicts for each nodule the malignancy as a regression value. To increase the amount of nodules, data augmentation is applied to the training samples. The classification network is trained using only the small scale of 3.2 cm as it was shown during preliminary experiments that this scale was able to learn the malignancy features better.

 \begin{figure*}
\centering
\includegraphics[width=1\textwidth]{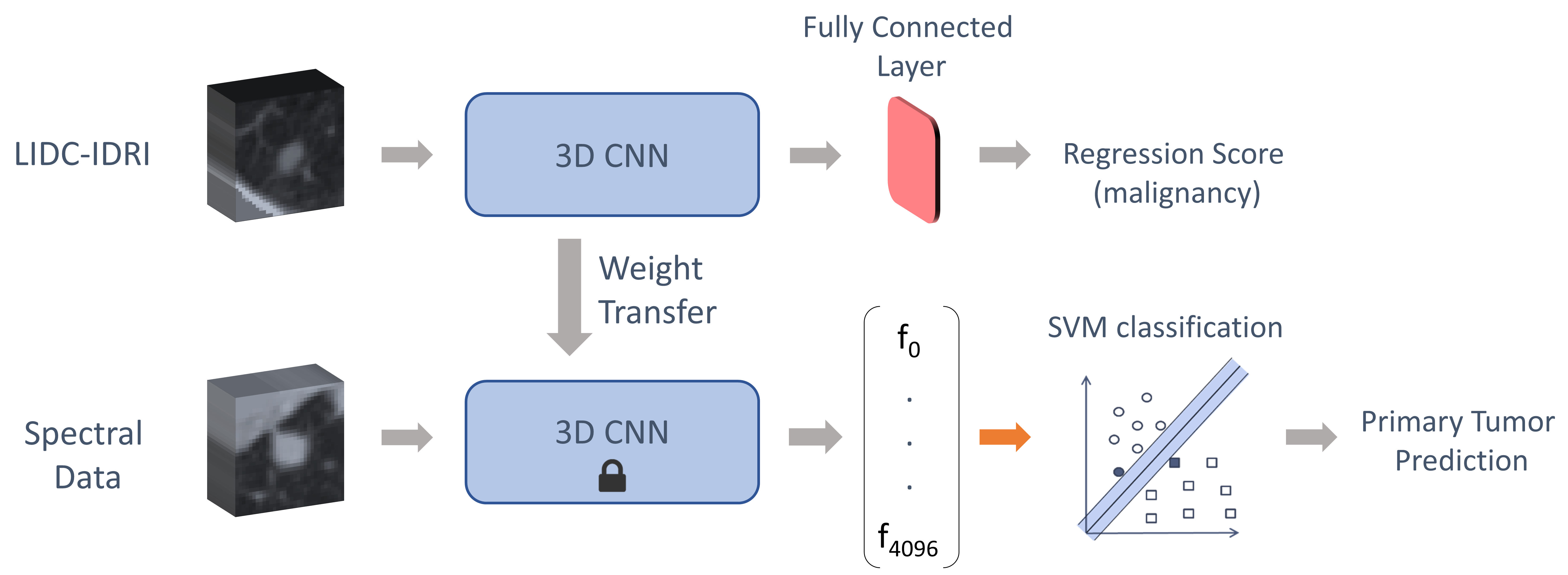}
\caption{Schematic overview of the classification method. In the pipeline at the top the CNN is trained using the LIDC-IDRI database, the 3D CNN corresponds to the box drawn onto the neural network in Figure \ref{fig:architecture}. The trained CNN is then transferred to act as feature extractor on the spectral data. The orange arrow indicates feature aggregation for scan-level predictions, whereas for nodule-level predictions the individual feature vectors are used directly. }
\label{fig:classOverview}
\end{figure*}

Once training on the LIDC-IDRI is completed, the convolutional layers of this network are extracted, containing the learned filter weights. Subsequently this network is used as a feature extractor for the spectral nodules, without any fine-tuning of the network. This results for each nodule in a feature vector of 4096 dimensions, which is then classified using SVM. A schematic overview of the proposed classification method is given in Figure \ref{fig:classOverview}. 

Classifying the feature vectors per nodule results in nodule-level classification. However, we are also interested in scan-level labeling. To obtain scan-level predictions the nodules per scan have to be combined into one feature vector per scan. As the number of nodules per scan differs, simple concatenation of the feature vectors is not possible. Two methods for combining the nodule features are applied in this study: element-wise combination of features and distance based similarity.  We also considered to use late fusion instead of concatenation, but initial tests did not show satisfactory results and it was therefore discarded. For the element-wise combination of features the maximum, minimum and mean are compared~\cite{gartner2002multi}.

The distance based metric is determined by defining for each nodule (feature vector) in scan $B_i$ the distance to the closest nodule of scan $B_j$. These minimum distances are then combined into one distance using either the maximum, minimum or mean. If each scan is represented by a bag $B_i=\{x_{ik}|k=1,...,n_i\}$ of $n_i$ feature vectors, the distance between bags $B_i$ and $B_j$ is given by \cite{cheplygina2015multiple}:
$$d(B_i,B_j) = \fun_k\min_l d(x_{ik},x_{jl}) $$
with \textit{func} the \textit{max}, \textit{min} or \textit{mean} and \textit{d} the euclidean distance.

For each scan the distance to every other scan is calculated, resulting per scan in a vector of length \textit{n scans}. These vectors are then used for SVM classification in dissimilarity space. 

Next, we want to observe whether spectral features can increase the performance of our classifier. In this study the used spectral features are obtained from a high (190 keV) and low (60 keV) mono-energetic view by applying the same feature extractor as used on the conventional scans to these images. The final feature vector is then obtained by concatenating the vectors of the different views. We also attempted to use the Compton scattering (CS) and photoelectric effect (PE) components as additional spectral representations, but this did not result in improved classification (see Appendix~\ref{appendix:results}).

\subsection{Network architecture}
An advantage of using a patch-based network for detection is that a similar network can be used for classification, needing only minor adaptations. The network architecture used in this study is a 3D CNN resembling a 3D VGG network and is shown in Figure~\ref{fig:architecture}~\citep{Tran2015,simonyan2014very}. The parameters used in the network are in line with commonly used parameters. During preliminary experiments the width and depth of the network were varied, both for the convolutional and fully connected layers, to choose the final architecture. The final architecture consists of four convolutional layer groups, each containing one or two convolutional layers, a max pool layer and batch normalization. The number of learned filters in the convolutional layers increases from 64 to 512, and the kernel size is 3x3x3. The max pool layers have a pool size of 2x2x2, with exception of the first max pool layer which only pools the x and y direction in order to equalize the size of the axes. 
After the convolutional layer groups one fully connected layer of 64 neurons is located, with an applied dropout ratio of 0.5. For the detection network this fully connected layer is followed by a sigmoid classification layer, yielding a probability, whereas for regression this layer is not present. Except for the sigmoid classification layer, the rectified linear unit (RELU) is used as activation function. 

The section of the network which is transferred as feature extractor for the spectral data consists of all convolutional and max pool layers, indicated by the dashed rectangle in Figure~\ref{fig:architecture}.

\section{Experiments} \label{experiments}
The CT scans used for the experiments in this study were first preprocessed, applying both intensity normalization and segmentation. In this section the preprocessing is described in more detail, followed by an explanation of the experimental setup of the detection and classification.   

\begin{figure*}
\centering
\includegraphics[width=1\textwidth]{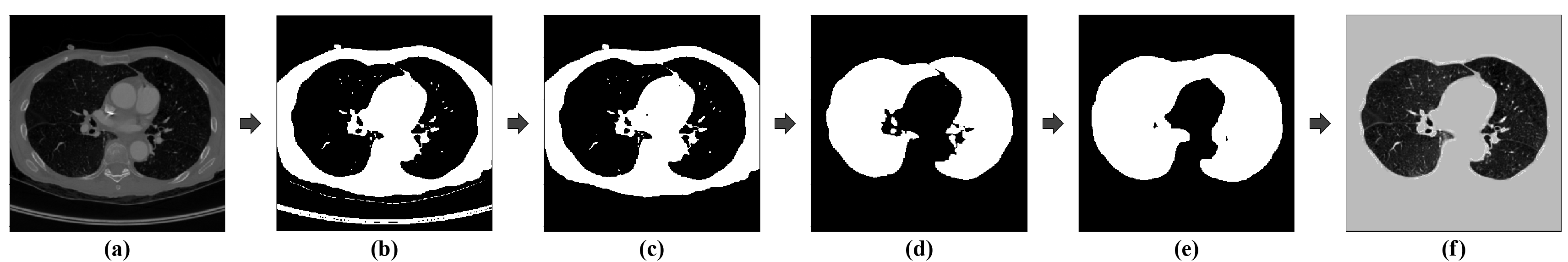}
\caption{Lung segmentation: (a) unprocessed CT image, (b) morphological closing with a kernel size of 3 mm and binarization with a threshold of -320 HU, (c) the largest connected volume above the threshold (the body) is selected and all other values are set to 0, (d) determination of background using the corner pixels, selecting the lungs as the largest volume below the threshold and inverting the image (e) dilation of the mask with a kernel size of 10 mm (f) final image after intensity normalization, the mask is applied to the image and all other values are set to 170, additionally all values larger than 210 in the space generated by the dilation are given 170 as well.}
\label{fig:segmentation}
\end{figure*}

\subsection{Preprocessing}
The raw data was first converted into Hounsfield units (HU), which is a measurement of relative densities obtained in CT. Due to different scan protocols, the resolution of different CT images varies. To make the data as homogeneous as possible, the images were rescaled to a resolution of 1 mm in the transversal plane, and 2 mm in the z-direction. The larger z resolution was chosen to reduce computational load, as usually the slice thickness is lower than the in-plane resolution. 

\paragraph{Lung mask extraction}
Thoracic CT scans contain other tissues next to the lungs. Because the detection algorithm can be distracted by surrounding tissue resembling nodules, it is desired to extract only the lungs from the images. The air in the lungs and the tissue surrounding the lungs have high contrast on a CT. For this reason a threshold based segmentation method can be used. The complete segmentation process which was applied to the scans is shown in Figure~\ref{fig:segmentation}. First, a morphological closing operation with a kernel size of 3 mm was applied to close potential narrow connections between the lungs and background due to noise. This was necessary to enable separation of lung tissue and background later on. A kernel size of 3 mm showed to give the best results during preliminary tests. Subsequently, the image was binarized using a threshold between the HUs of lung and tissue. The exact threshold does not have a significant effect on the results of this method, but was chosen at -320 HU~\citep{preprocessing}. To remove any background objects the patient's body was selected as the largest connect component above the threshold, and all other values were set to 0 (Figure~\ref{fig:segmentation}c). Next, all pixels connected to the corner pixels were defined as background. The final lung mask was then obtained by selecting the largest connected volume below the threshold (the lungs) and inverting the image (Figure \ref{fig:segmentation}d).
Before applying the lung mask to the image, it was dilated with a kernel size of 10 mm to obtain not solely the lungs but also part of the lung walls as these are considered important for detection and classification (Figure \ref{fig:segmentation}e). Furthermore, nodules can stick to the lung wall resulting in under-segmentation of lung tissue when using the mask directly. All segmentations were checked by hand, and in some cases adaptations had to be made to the kernel size of the closing operation to correctly segment both lungs. In the spectral dataset, some scans included the head or had a tracheotomy, resulting in no separation of background and lung. In these cases the head was manually removed from the scan by setting these slices to zero. For some patients in the spectral dataset it was not possible to obtain correct segmentation using this method. Because for classification the segmentation was not expected to be of great importance, no segmentation was applied to these scans. A list of the adapted parameters and missing segmentations for the spectral dataset can be found in Appendix~\ref{appendix:segmentations}.

\paragraph{Intensity normalization}
The intensities were  clipped between -1200 and 600 HU as values outside this range were considered noise or irrelevant for this study~\citep{igrt}. Next the values were linearly rescaled between 0 and 255 (8-bit representation). Then the image was multiplied with the extracted lung mask, where all values outside the masked area were given a value of 170. The luminance of 170 corresponds to a HU value of 0, which it the HU of water~\citep{igrt}. Furthermore, all values larger than 210 in the space generated by the dilation of the lung mask were set to water value as well. The reason for this is that the bones in this area, having high CT values, could be mislabeled as calcified nodules~\citep{igrt}. The chosen luminance level of 210 corresponds to a HU value of 300 and is adopted from~\citep{igrt}. The value is higher than most tissue values, but below the values from bony areas. The final normalized and segmented image is shown in Figure~\ref{fig:segmentation}f.

\subsection{Nodule detection}
The detection network was trained on the LIDC-IDRI database and later applied to the conventional multi-energetic representation of the spectral dataset. For development the LIDC-IDRI was split up into three subsets; the training (70\%), validation (10\%) and testing set (20\%). The testing set was not seen during training and optimization, and only used during final evaluation. As described earlier, the network was trained and evaluated using scales of 3.2 and 6.4 cm. For the LIDC-IDRI database, all nodules have a diameter between 0.3 and 3 cm. The small scale thus realizes the capture of enough details, mostly for small nodules in the dataset, whereas the large scale ensures that enough background is taken into account, which can be important in areas like the bronchi.

\paragraph{Training sample selection}
During training each batch consisted of an equal amount of positive and negative samples in order to obtain balanced training of the network. A positive sample is defined as a sample containing a complete nodule, whereas a negative sample has no overlap with a nodule at all. 

Positive samples were obtained by cropping cubes around each nodule annotation in the images. The samples were augmented using translation and all 3D flips. Translation was necessary because during sliding window evaluation of the whole image, nodules are not necessarily centered in a patch. It was implemented by cropping cubes around the nodules with a random shift. This shift was limited to 4 mm from the center point in each direction, which corresponds to the size of the cube to which the prediction of the patch is assigned during evaluation (8 mm). As the cube in which the center of the nodule is present should predict the highest probability, the training shift was limited to this value. Furthermore, it was ensured that the whole nodule was always completely contained in the cube. Scaling and rotation were also considered, however these did not yield any improvement on the validation set and where therefore not adopted.  

Negative samples were obtained in three ways. Firstly, random negative samples were cropped from the image, of which the center point of the sample was located inside the lung segmentation. Random sampling ensured that all parts of the image were represented in the samples. Secondly, negative samples were selected using the LUNA2016 candidate list, containing nodule candidate locations predicted by classical nodule detection algorithms~\citep{LUNA2016}. As the LUNA2016 candidate list contained more difficult  examples (resembling nodules) than random sampling, twice as many  samples from the candidate list (40/scan) were used as random samples (20/scan). The last type of negative samples were obtained by hard negative mining. Hard negative mining is the construction of negative samples from false positive locations after initial training. These samples were created by thresholding the predicted probabilities in the training CT scans, and using the false positive locations as new samples. For each scan a maximum of 10 false positive samples was added to the dataset, which showed to create enough challenging examples during training.

\paragraph{Prediction validation} \label{section:evaluation}
The generated predictions for the LIDC-IDRI were validated using the available annotations, and the list of irrelevant findings from the LUNA2016 challenge. In order to determine the sensitivity and the number of false positives, the predicted probabilities were binarized and clustered using connected component labeling~\citep{wu2005optimizing}. Each annotated nodule which overlapped with a predicted cluster was considered a correct detection, whereas each predicted cluster not overlapping with either a nodule or irrelevant finding was a false positive (FP). An irrelevant finding was thus not counted as true detection nor as false positive. It should be noted that during hard negative mining in sample selection no irrelevant findings were used during evaluation. The final sensitivity and false positives were calculated for a range of thresholds, resulting in a free receiver operator curve (FROC). The analysis was performed for both the validation and testing set. 

The predictions for the spectral dataset were validated using the available annotations. Nodules $\leq$ 3 mm or $\geq$ 3 cm were excluded from the annotations to be consistent with the annotations in the training data.

\subsection{Nodule classification}
An overview of the nodule classification is shown in Figure~\ref{fig:classOverview}. This section will explain how the regression model was applied to the LIDC-IDRI dataset, and the subsequent use as feature extractor on the spectral dataset.

\paragraph{LIDC-IDRI regression}
Each of the nodules in the LIDC-IDRI was scored by multiple radiologists. In order to obtain one score per nodule, the ratings of the observers were averaged. The input samples were obtained by cropping cubes from the scans, each containing a centered nodule. Because all nodules in the LIDC-IDRI are smaller than 3 cm, each nodule was able to fit completely in the input cube using only the small scale of 3.2 cm. As data augmentation all 3D flips, random rotation and scaling between 0.8 and 1.2 were applied, which showed to give the best results during initial experiments. Scaling with larger factors was also applied (in a range of 0.5-1.5), but this did not improve the results and was thus not adopted. The network parameters were fine-tuned using a train and validation set, and final evaluation was performed using 10-fold cross validation.   

The performance is reported as both the mean absolute error (MAE) between the predicted and actual value, and as the one-off-accuracy. This regression accuracy is adapted from~\citep{shen2016learning} and is defined as the percentage of predictions with an error smaller than 1. This constraint accounts for some of the inter-observer variability, as the observers seldom agreed on one malignancy score.

\paragraph{Feature extraction and SVM classification}
The trained regression network is used to extract a feature vector for each nodule in the spectral dataset. These features are then either classified separately, yielding nodule-level predictions, or aggregated to scan-level features, resulting in scan-level predictions. Before classification the (aggregated) vectors were normalized to unit length. The classification of the features was performed using linear SVM, in which $C \in [0.01,0.03,0.1,0.3,1,3,10,30,100,300,1000]$ was determined using optimization of the $F_1$-macro score in 10-fold cross-validation. The $F_1$-macro is defined as the unweighted mean of the $F_1$ score per class: $$ F_1-macro= \dfrac{1}{n} \sum_{k=1}^{n}  2  \cdot \dfrac{precision_k \cdot recall_k}{precision_k + recall_k}, $$ with \textit{n} the number of classes. In this study the $F_1$-macro is abbreviated as the F-score.

The available data contained five groups: benign multinodular, benign, primary lung cancer, colorectal cancer and melanoma. Because preliminary results suggested that with the current features it was not possible to differentiate between melanoma and colorectal cancer (see Appendix~\ref{appendix:results}), we decided to classify these together as metastases. Furthermore, we decided to classify benign and benign multinodular together as benign, because it is expected that these two groups will have similar appearance per nodule, and a differentiation between those two groups can be readily made knowing the number of nodules. Final experiments were performed using both two- and three-class labels. In the three-class problem the nodules were classified as benign, primary lung cancer and metastases. For the two-class problem, primary lung cancer and metastases were classified together as malignant, resulting in a classification of benign versus malignant. All experiments were performed for the conventional images, and for the combination of conventional and spectral images.   

To evaluate the classification both F-score and accuracy were used. The accuracy is the total number of nodules which is correctly classified. For each of the individual classification scores a permutation test was performed. This is a test in which the labels are permuted multiple times, indicating whether the obtained scores are significantly different from scores obtained by chance~\citep{ojala2010permutation}. Furthermore, for the conventional versus the spectral representations a paired t-test was done for the different folds of the 10-fold cross-validation.

\subsection{Implementation}
The networks were trained and evaluated using the Titan XP GPU with 12 GB memory from Nvidia. The total batch size was 40 for the smaller crops of 3.2 cm, and 10 for the larger crops due to memory constraints of the GPU. During training the Adam optimization algorithm was used with an initial learning rate of 0.0001~\cite{kinga2015method}. The network was trained using Tensorflow and Keras. The LinearSVM function from scikit-learn was used with `hinge' loss \citep{scikit-learn}. During preliminary experiments `hinge' loss (default of another SVM implementation in scikit-learn) showed better results than the default `squared hinge' loss and was therefore adopted. The multi-class classification is implemented by LinearSVM using a one-vs-rest implementation.

\section{Results} \label{results}
In this section the obtained results are presented. First the detection performance on both the LIDC-IDRI and the spectral dataset is shown, followed by the classification results. 

\subsection{Detection}
\begin{figure*}
	\centering
	\setlength\figureheight{7.5 cm}
	\setlength\figurewidth{0.48\textwidth}
	\begin{minipage}[b]{0.48\textwidth}
    	\centering
\begin{tikzpicture}
\def\colorModel{RGB}
\clip (-1.5,-1.1) rectangle (8,6.2);
\definecolor{color2}{RGB}{166,206,227}
\definecolor{color3}{RGB}{178,223,138}
\definecolor{color4}{RGB}{251,154,153}
\definecolor{color10}{RGB}{31,120,180}
\definecolor{color1}{RGB}{51,160,44}
\definecolor{color0}{RGB}{227,26,28}
\begin{axis}[
height=\figureheight,
legend cell align={left},
legend entries={{Validation: 3.2 cm},{Validation: 6.4 cm},{Validation: Both Scales},{Testing: 3.2 cm},{Testing: 6.4 cm},{Testing: Both Scales}},
legend style={at={(0.97,0.03)}, anchor=south east, draw=white!80.0!black},
tick align=outside,
tick pos=left,
width=\figurewidth,
x grid style={white!69.01960784313725!black},
xlabel={average FP/scan},
xmin=0.125, xmax=16,
xtick={0.125,2,4,8,12,16},
xticklabels={0.125,2,4,8,12,16},
y grid style={white!69.01960784313725!black},
ylabel={sensitivity},
ymin=0.4, ymax=1,
ytick={0.0,0.2,0.4,0.6,0.8,1},
yticklabels={0.0,0.2,0.4,0.6,0.8,1.0}
]
\addlegendimage{ line width=1.5 pt, mark=square*,mark size=0.7,color2}
\addlegendimage{ line width=1.5 pt, mark=square*,mark size=0.7, color3}
\addlegendimage{line width=1.5 pt, mark=square*,mark size=0.7, color4}
\addlegendimage{line width=1.5 pt,mark=*,mark size=0.7, color10}
\addlegendimage{line width=1.5 pt,mark=*,mark size=0.7, color1}
\addlegendimage{line width=1.5 pt,mark=*,mark size=0.7,color0}
\addplot [ thick, color2,  mark= square*, mark size=0.7, mark options={solid}]
table [row sep=\\]{%
190.188118811881	0.99290780141844 \\
146.524752475248	0.99290780141844 \\
118.891089108911	0.99290780141844 \\
100.089108910891	0.99290780141844 \\
93.0792079207921	0.99290780141844 \\
86.6732673267327	0.99290780141844 \\
51.3564356435644	0.99290780141844 \\
18.5148514851485	0.99290780141844 \\
15.2970297029703	0.985815602836879 \\
12.4455445544554	0.985815602836879 \\
7.94059405940594	0.985815602836879 \\
5.72277227722772	0.971631205673759 \\
4.63366336633663	0.964539007092199 \\
3.69306930693069	0.964539007092199 \\
3.10891089108911	0.957446808510638 \\
2.69306930693069	0.936170212765957 \\
2.26732673267327	0.914893617021277 \\
1.4950495049505	0.879432624113475 \\
1.24752475247525	0.843971631205674 \\
0.97029702970297	0.801418439716312 \\
0.732673267326733	0.794326241134752 \\
0.574257425742574	0.75886524822695 \\
0.158415841584158	0.546099290780142 \\
0	0 \\
};
\addplot [ thick, color3, mark=square*,  mark size=0.7, mark options={solid}]
table [row sep=\\]{%
19.8316831683168	0.971631205673759 \\
13.5544554455446	0.957446808510638 \\
10.8712871287129	0.950354609929078 \\
8.82178217821782	0.950354609929078 \\
8.21782178217822	0.950354609929078 \\
7.6039603960396	0.950354609929078 \\
5.02970297029703	0.943262411347518 \\
3.87128712871287	0.943262411347518 \\
3.27722772277228	0.943262411347518 \\
2.71287128712871	0.936170212765957 \\
1.85148514851485	0.907801418439716 \\
1.53465346534653	0.900709219858156 \\
1.37623762376238	0.879432624113475 \\
1.03960396039604	0.843971631205674 \\
0.841584158415842	0.822695035460993 \\
0.752475247524752	0.773049645390071 \\
0.534653465346535	0.730496453900709 \\
0.257425742574257	0.531914893617021 \\
0.198019801980198	0.375886524822695 \\
0.0693069306930693	0.184397163120567 \\
0.0495049504950495	0.127659574468085 \\
0.0297029702970297	0.0851063829787234 \\
0	0 \\
};
\addplot [thick, color4,   mark=square*, mark size=0.7, mark options={solid}]
table [row sep=\\]{%
150.148514851485	1 \\
103.376237623762	1 \\
80.029702970297	1 \\
64.1782178217822	1 \\
58.4752475247525	1 \\
53.2079207920792	1 \\
16.7029702970297	0.99290780141844 \\
11.3465346534653	0.99290780141844 \\
6.79207920792079	0.978723404255319 \\
2.9009900990099	0.957446808510638 \\
1.54455445544554	0.929078014184397 \\
1.20792079207921	0.921985815602837 \\
1.01980198019802	0.893617021276596 \\
0.811881188118812	0.886524822695035 \\
0.653465346534653	0.872340425531915 \\
0.554455445544555	0.836879432624113 \\
0.415841584158416	0.75886524822695 \\
0.277227722772277	0.645390070921986 \\
0.168316831683168	0.51063829787234 \\
0.099009900990099	0.290780141843972 \\
0.0594059405940594	0.170212765957447 \\
0.0396039603960396	0.120567375886525 \\
0	0 \\
};
\addplot [thick, color10, mark=*, mark size=0.7, mark options={solid}]
table [row sep=\\]{%
190.758620689655	0.972093023255814 \\
149.068965517241	0.962790697674419 \\
124.109195402299	0.962790697674419 \\
106.718390804598	0.962790697674419 \\
99.3908045977011	0.962790697674419 \\
92.7011494252874	0.962790697674419 \\
51.9770114942529	0.958139534883721 \\
18.4827586206897	0.953488372093023 \\
15.4597701149425	0.948837209302326 \\
12.6149425287356	0.944186046511628 \\
7.88505747126437	0.934883720930233 \\
5.67241379310345	0.916279069767442 \\
4.67241379310345	0.911627906976744 \\
3.60919540229885	0.906976744186046 \\
3.08045977011494	0.893023255813953 \\
2.6264367816092	0.893023255813953 \\
2.2183908045977	0.87906976744186 \\
1.61494252873563	0.818604651162791 \\
1.36206896551724	0.790697674418605 \\
1.16666666666667	0.772093023255814 \\
1.01724137931034	0.758139534883721 \\
0.873563218390805	0.725581395348837 \\
0.189655172413793	0.423255813953488 \\
0					0				\\
};
\addplot [ thick, color1,  mark=*,  mark size=0.7, mark options={solid}]
table [row sep=\\]{%
19.1034482758621	0.981395348837209 \\
12.7126436781609	0.967441860465116 \\
9.83333333333333	0.953488372093023 \\
8.09770114942529	0.953488372093023 \\
7.48850574712644	0.948837209302326 \\
7.05747126436782	0.948837209302326 \\
4.52873563218391	0.934883720930233 \\
3.52298850574713	0.930232558139535 \\
2.91379310344828	0.911627906976744 \\
2.48850574712644	0.911627906976744 \\
1.81034482758621	0.906976744186046 \\
1.51149425287356	0.87906976744186 \\
1.27586206896552	0.851162790697674 \\
1.0632183908046	0.827906976744186 \\
0.867816091954023	0.809302325581395 \\
0.71264367816092	0.786046511627907 \\
0.540229885057471	0.73953488372093 \\
0.247126436781609	0.581395348837209 \\
0.17816091954023	0.427906976744186 \\
0.0862068965517241	0.274418604651163 \\
0.028735632183908	0.172093023255814 \\
0.0114942528735632	0.102325581395349 \\
0	0 \\
};
\addplot [ thick, color0,  mark=*,  mark size=0.7, mark options={solid}]
table [row sep=\\]{%
152.155172413793	0.990697674418605 \\
109.155172413793	0.976744186046512 \\
84.2011494252874	0.976744186046512 \\
67.3275862068966	0.976744186046512 \\
59.8735632183908	0.972093023255814 \\
53.5114942528736	0.972093023255814 \\
16.5057471264368	0.967441860465116 \\
11.0977011494253	0.958139534883721 \\
6.47701149425287	0.958139534883721 \\
2.81609195402299	0.93953488372093 \\
1.5632183908046	0.897674418604651 \\
1.36206896551724	0.888372093023256 \\
1.18390804597701	0.869767441860465 \\
0.994252873563218	0.837209302325581 \\
0.890804597701149	0.823255813953488 \\
0.758620689655172	0.8 \\
0.591954022988506	0.758139534883721 \\
0.310344827586207	0.637209302325581 \\
0.218390804597701	0.53953488372093 \\
0.103448275862069	0.32093023255814 \\
0.0689655172413793	0.246511627906977 \\
0.0172413793103448	0.144186046511628 \\
0	0 \\
};    
\end{axis}
\end{tikzpicture}
	\end{minipage}
    \begin{minipage}[b]{0.48\textwidth}
   		\centering
\begin{tikzpicture}
\clip (-1.5,-1.1) rectangle (8,6.2);

\definecolor{color2}{RGB}{166,206,227}
\definecolor{color3}{RGB}{178,223,138}
\definecolor{color4}{RGB}{251,154,153}
\definecolor{color10}{RGB}{31,120,180}
\definecolor{color1}{RGB}{51,160,44}
\definecolor{color0}{RGB}{227,26,28}
\begin{axis}[
height=\figureheight,
legend cell align={left},
legend entries={{Validation: 3.2 cm},{Validation: 6.4 cm},{Validation: Both Scales},{Testing: 3.2 cm},{Testing: 6.4 cm},{Testing: Both Scales}},
legend style={at={(0.97,0.03)}, anchor=south east, draw=white!80.0!black},
tick align=outside,
tick pos=left,
width=\figurewidth,
x grid style={white!69.01960784313725!black},
xlabel={average FP/scan},
xmin=0.125, xmax=4,
xtick={0.125,1,2,3,4},
xticklabels={0.125,1,2,3,4},
y grid style={white!69.01960784313725!black},
ylabel={sensitivity},
ymin=0.5, ymax=1,
ytick={0.5,0.6,0.7,0.8,0.9,1.0},
yticklabels={0.5,0.6,0.7,0.8,0.9,1.0}
]
\addlegendimage{ line width=1.5 pt, mark=square*,mark size=0.7,color2}
\addlegendimage{ line width=1.5 pt, mark=square*,mark size=0.7, color3}
\addlegendimage{line width=1.5 pt, mark=square*,mark size=0.7, color4}
\addlegendimage{line width=1.5 pt,mark=*,mark size=0.7, color10}
\addlegendimage{line width=1.5 pt,mark=*,mark size=0.7, color1}
\addlegendimage{line width=1.5 pt,mark=*,mark size=0.7,color0}
\addplot [thick, color2, mark= square*, mark size=0.7, mark options={solid}]
table [row sep=\\]{%
190.188118811881	0.99290780141844 \\
146.524752475248	0.99290780141844 \\
118.891089108911	0.99290780141844 \\
100.089108910891	0.99290780141844 \\
93.0792079207921	0.99290780141844 \\
86.6732673267327	0.99290780141844 \\
51.3564356435644	0.99290780141844 \\
18.5148514851485	0.99290780141844 \\
15.2970297029703	0.985815602836879 \\
12.4455445544554	0.985815602836879 \\
7.94059405940594	0.985815602836879 \\
5.72277227722772	0.971631205673759 \\
4.63366336633663	0.964539007092199 \\
3.69306930693069	0.964539007092199 \\
3.10891089108911	0.957446808510638 \\
2.69306930693069	0.936170212765957 \\
2.26732673267327	0.914893617021277 \\
1.4950495049505	0.879432624113475 \\
1.24752475247525	0.843971631205674 \\
0.97029702970297	0.801418439716312 \\
0.732673267326733	0.794326241134752 \\
0.574257425742574	0.75886524822695 \\
0.158415841584158	0.546099290780142 \\
0	0\\
};
\addplot [thick, color3,  mark=square*,  mark size=0.7, mark options={solid}]
table [row sep=\\]{%
19.8316831683168	0.971631205673759 \\
13.5544554455446	0.957446808510638 \\
10.8712871287129	0.950354609929078 \\
8.82178217821782	0.950354609929078 \\
8.21782178217822	0.950354609929078 \\
7.6039603960396	0.950354609929078 \\
5.02970297029703	0.943262411347518 \\
3.87128712871287	0.943262411347518 \\
3.27722772277228	0.943262411347518 \\
2.71287128712871	0.936170212765957 \\
1.85148514851485	0.907801418439716 \\
1.53465346534653	0.900709219858156 \\
1.37623762376238	0.879432624113475 \\
1.03960396039604	0.843971631205674 \\
0.841584158415842	0.822695035460993 \\
0.752475247524752	0.773049645390071 \\
0.534653465346535	0.730496453900709 \\
0.257425742574257	0.531914893617021 \\
0.198019801980198	0.375886524822695 \\
0.0693069306930693	0.184397163120567 \\
0.0495049504950495	0.127659574468085 \\
0.0297029702970297	0.0851063829787234 \\
0	0 \\
};
\addplot [thick, color4,   mark=square*, mark size=0.7, mark options={solid}]
table [row sep=\\]{%
150.148514851485	1 \\
103.376237623762	1 \\
80.029702970297	1 \\
64.1782178217822	1 \\
58.4752475247525	1 \\
53.2079207920792	1 \\
16.7029702970297	0.99290780141844 \\
11.3465346534653	0.99290780141844 \\
6.79207920792079	0.978723404255319 \\
2.9009900990099	0.957446808510638 \\
1.54455445544554	0.929078014184397 \\
1.20792079207921	0.921985815602837 \\
1.01980198019802	0.893617021276596 \\
0.811881188118812	0.886524822695035 \\
0.653465346534653	0.872340425531915 \\
0.554455445544555	0.836879432624113 \\
0.415841584158416	0.75886524822695 \\
0.277227722772277	0.645390070921986 \\
0.168316831683168	0.51063829787234 \\
0.099009900990099	0.290780141843972 \\
0.0594059405940594	0.170212765957447 \\
0.0396039603960396	0.120567375886525 \\
0	0 \\
};
\addplot [thick, color10, mark=*, mark size=0.7, mark options={solid}]
table [row sep=\\]{%
190.758620689655	0.972093023255814 \\
149.068965517241	0.962790697674419 \\
124.109195402299	0.962790697674419 \\
106.718390804598	0.962790697674419 \\
99.3908045977011	0.962790697674419 \\
92.7011494252874	0.962790697674419 \\
51.9770114942529	0.958139534883721 \\
18.4827586206897	0.953488372093023 \\
15.4597701149425	0.948837209302326 \\
12.6149425287356	0.944186046511628 \\
7.88505747126437	0.934883720930233 \\
5.67241379310345	0.916279069767442 \\
4.67241379310345	0.911627906976744 \\
3.60919540229885	0.906976744186046 \\
3.08045977011494	0.893023255813953 \\
2.6264367816092	0.893023255813953 \\
2.2183908045977	0.87906976744186 \\
1.61494252873563	0.818604651162791 \\
1.36206896551724	0.790697674418605 \\
1.16666666666667	0.772093023255814 \\
1.01724137931034	0.758139534883721 \\
0.873563218390805	0.725581395348837 \\
0.189655172413793	0.423255813953488 \\
0	0\\
};
\addplot [thick, color1,  mark=*,  mark size=0.7, mark options={solid}]
table [row sep=\\]{%
19.1034482758621	0.981395348837209 \\
12.7126436781609	0.967441860465116 \\
9.83333333333333	0.953488372093023 \\
8.09770114942529	0.953488372093023 \\
7.48850574712644	0.948837209302326 \\
7.05747126436782	0.948837209302326 \\
4.52873563218391	0.934883720930233 \\
3.52298850574713	0.930232558139535 \\
2.91379310344828	0.911627906976744 \\
2.48850574712644	0.911627906976744 \\
1.81034482758621	0.906976744186046 \\
1.51149425287356	0.87906976744186 \\
1.27586206896552	0.851162790697674 \\
1.0632183908046	0.827906976744186 \\
0.867816091954023	0.809302325581395 \\
0.71264367816092	0.786046511627907 \\
0.540229885057471	0.73953488372093 \\
0.247126436781609	0.581395348837209 \\
0.17816091954023	0.427906976744186 \\
0.0862068965517241	0.274418604651163 \\
0.028735632183908	0.172093023255814 \\
0.0114942528735632	0.102325581395349 \\
0	0 \\
};
\addplot [thick, color0,  mark=*,  mark size=0.7, mark options={solid}]
table [row sep=\\]{%
152.155172413793	0.990697674418605 \\
109.155172413793	0.976744186046512 \\
84.2011494252874	0.976744186046512 \\
67.3275862068966	0.976744186046512 \\
59.8735632183908	0.972093023255814 \\
53.5114942528736	0.972093023255814 \\
16.5057471264368	0.967441860465116 \\
11.0977011494253	0.958139534883721 \\
6.47701149425287	0.958139534883721 \\
2.81609195402299	0.93953488372093 \\
1.5632183908046	0.897674418604651 \\
1.36206896551724	0.888372093023256 \\
1.18390804597701	0.869767441860465 \\
0.994252873563218	0.837209302325581 \\
0.890804597701149	0.823255813953488 \\
0.758620689655172	0.8 \\
0.591954022988506	0.758139534883721 \\
0.310344827586207	0.637209302325581 \\
0.218390804597701	0.53953488372093 \\
0.103448275862069	0.32093023255814 \\
0.0689655172413793	0.246511627906977 \\
0.0172413793103448	0.144186046511628 \\
0	0 \\
};
\end{axis}
\end{tikzpicture}
    \end{minipage}
	\caption{FROC curves of the network performance on the LIDC-IDRI database. The performance is shown both on the validation and testing set for all scales. On the right a zoomed version of the graph is shown for better visibility of this area.}
\label{fig:froclidc}
\end{figure*}
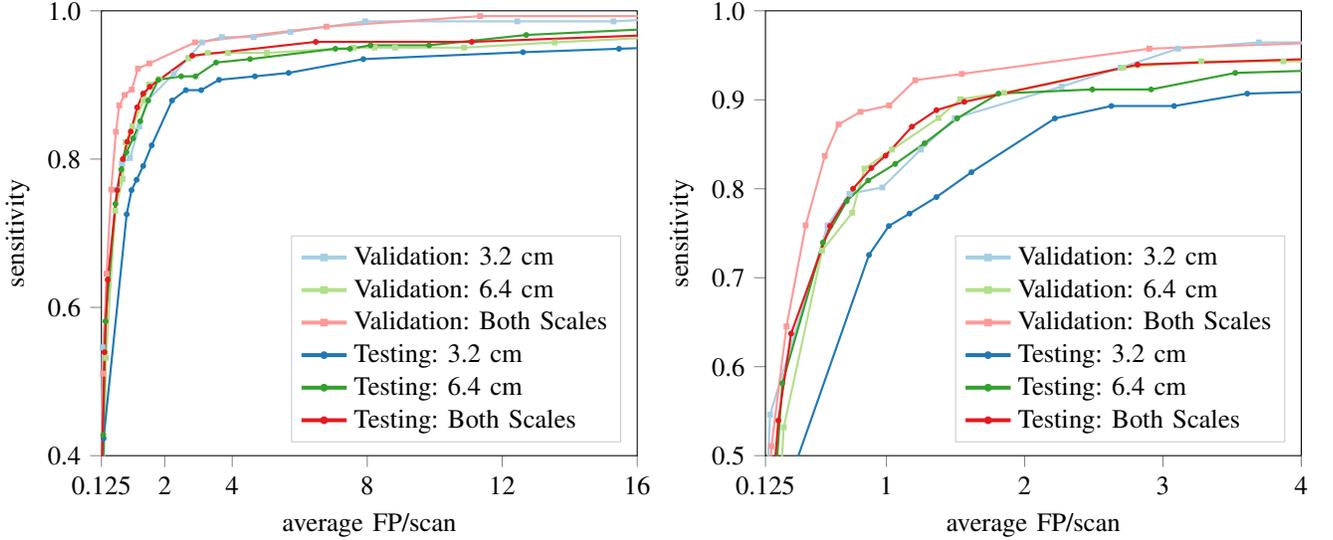

In Figure~\ref{fig:froclidc} the FROC curves from the detection performance on the LIDC-IDRI database are shown for both the validation and testing subset. On the left side the curve is shown for the entire FP range and on the right only the low FP rates are shown. From the FROC curves on both subsets it can be observed that the performance on the testing set is lower than on the validation set. Figure~\ref{fig:froclidc} also shows that the combination of both the 3.2 and 6.4 cm scale provides the best performance as opposed to just using either one of the individual scales. We consider the performance of the combined scales as the final performance of our model. In Table~\ref{table:detection} these results are summarized along with those of the top four performing teams from the LUNA2016 challenge as a means of comparison. The average sensitivity we achieved would have obtained the fourth position in the challenge. However, this score is lowered considerably by our low sensitivity at very small FP rates. At a higher FP rate of 8 FP/scan our method performs significantly better than the top performing team from LUNA2016. In Appendix \ref{appendix:results} a figure of the complete FROC curves of the different methods is shown. 

The trained detection network was applied to the spectral dataset to obtain insight in the generalization ability of the network. The performance on this dataset is depicted in Table~\ref{table:detection} as well, showing that the average sensitivity is only 0.288. Especially at low FP rates the sensitivity on this dataset is almost zero. The FROC curves from this dataset can be found in Appendix \ref{appendix:results}.

\renewcommand{\arraystretch}{1.2}
\begin{table}\centering
\caption{Detection performance from our method and the top four LUNA2016 contestants. The sensitivities from the contestants at certain FP rates were extracted from a figure containing the FROC curves and are therefore shown with less precision~\cite{LUNA2016}. The average sensitivity was calculated as an average sensitivity at 7 FP rates: 0.125, 0.25, 0.5, 1, 2, 4, 8.}
\begin{tabular}{@{}lrrrr@{}} \toprule
 \multicolumn{1}{c}{} &  \multicolumn{3}{c}{Sensitivity (x100)} \\
\cmidrule{2-4} 
Method & averaged  & 0.125 FP/scan &  1 FP/scan & 8 FP/scan \\ \midrule
\citet{Znet} & 81.1    & 	66 &  83 & 91 \\
\citet{aidence} & 80.7   & 60 & 85&  91           \\ 
\citet{jianpei} & 77.6  & 62 & 80 & 86\\
\citet{lopez2015large} & 74.2 & 60 & 76 &  84  \\
 \addlinespace
Proposed CNN \\
\hspace{3mm}LIDC-IDRI &  75.8 & 36.1 & 83.8 & 95.8\\
\hspace{3mm}Spectral Data & 28.8 & 1.5 & 17.9 &76.1\\
\bottomrule
\end{tabular}
\label{table:detection}
\end{table}

\subsection{Classification}
The results of the malignancy regression on the LIDC-IDRI database are shown in Table \ref{table:LIDC-IDRI malignancy}. For a frame of reference Table~\ref{table:LIDC-IDRI malignancy} lists the MAE and one-off-accuracy for our own methodology as well as for several other studies that applied malignancy regression on the LIDC-IDRI. Our proposed CNN achieved a one-off-accuracy of 91.22\% and a MAE of 0.448 which outperforms~\cite{shen2016learning} and~\cite{buty2016characterization}, and performs approximately equal to~\cite{Hussein2017}. We also tried to compare our performance to~\cite{sahu2018lightweight}, but because they apply logistic regression all errors are rounded before averaging. This makes their results not directly comparable to our performance. 

\renewcommand{\arraystretch}{1.2}
\begin{table}\centering
\caption{Malignancy estimation on LIDC-IDRI. The results of our own method, and of all papers except for~\citep{shen2016learning} were obtained using 10-fold cross-validation. Some entries are empty as this information was not available.}
\begin{tabular*}{\linewidth}{@{\extracolsep{\fill}}lrr@{}} \toprule
Method & MAE & One-off-accuracy (\%)   \\ \midrule
\citet{shen2016learning} & -    & 90.99 \\
\citet{Hussein2017} & 0.459   & 91.26            \\ 
\citet{buty2016characterization} & - & 82.4 \\
\addlinespace
Proposed regression CNN  & 0.448 (0.03)  & 91.22 (1.69)       \\
\bottomrule
\end{tabular*}
\label{table:LIDC-IDRI malignancy}
\end{table}

The trained CNN was used as a feature extractor on the spectral dataset. These features were subsequently classified using SVM, both per nodule and per scan. To perform scan-level classification we tested several aggregation methods using only the conventional views of the spectral dataset. The obtained performance for each method can be found in Table~\ref{table:aggregation}. Element-wise maximum provided the best average results for the two- and three-class problems. Therefore, we used this aggregation function in the rest of the experiments to obtain scan-level predictions.

\renewcommand{\arraystretch}{1.2}
\begin{table}\centering
\caption{Comparison of the F-score (x100) for the nodule aggregation methods for benign/malignant (Ben/Mal) and benign/primary lung/metastases (Ben/L/Met).}
\begin{tabular*}{\linewidth}{@{\extracolsep{\fill}}lrrrcrrr@{}} \toprule
\multicolumn{1}{c}{} & \multicolumn{3}{c}{Element-wise} & & \multicolumn{3}{c}{Distances} \\
\cmidrule{2-4} \cmidrule{6-8} 
 & max & min & mean && maxmin & minmin & meanmin \\ \midrule 
Ben/Mal & 77.8 & 37.1  &75.1 && 75.1 & 78.7 &72.5  \\
Ben/L/Met & 67.8 & 59.8 &25.1   && 56.1 & 57.2 &51.5\\
\bottomrule
\end{tabular*}
\label{table:aggregation}
\end{table}

\begin{figure*}
	\centering
	\setlength\figureheight{6.5cm}
	\setlength\figurewidth{0.48\textwidth}
	\includegraphics{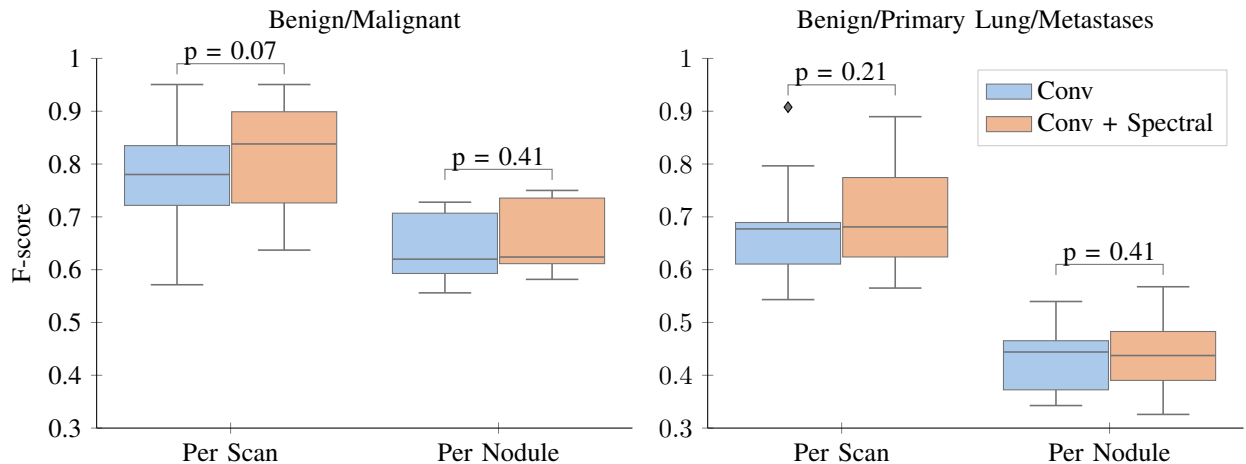}
	\caption{Box plots of the F-scores obtained by 10-fold cross-validation. Results are shown both for scan-level and nodule-level predictions, and for both the conventional features and a combination of conventional and spectral features.}
\label{fig:BoxplotsF1macro}
\end{figure*}

For both the two- and three-class problem, classification was performed using 10-fold cross-validation. The F-scores resulting from these experiments are shown in box plots in Figure~\ref{fig:BoxplotsF1macro}. The spectral features used in these experiments are from a low and high mono-energetic representation. In Figure~\ref{fig:BoxplotsF1macro} p-values are indicated which were obtained from a paired t-test between the different folds of the cross-validation. All p-values are higher than 0.05, indicating that no significant differences exists between the performance of the conventional features and of the combination of spectral and conventional features. The same graph depicting the resulting accuracy can be found in Appendix~\ref{appendix:results}. In Table~\ref{table:ClassificationResults} an overview of the resulting accuracies and F-scores for all experiments is presented. All individual classification scores in this table are statistically significant using a permutation test (p$<0$.01).

\renewcommand{\arraystretch}{1.2}
\begin{table}\centering
\caption{Classification results (x100) for benign/malignant (Ben/Mal) and benign/primary lung/metastases (Ben/L/Met). All individual metrics are significant (p$<$0.01) using permutation testing. The values between brackets are the standard deviations. }
\begin{tabular}{@{}lrrcrr@{}} \toprule
& \multicolumn{2}{c}{Conventional} & \phantom{ab}& \multicolumn{2}{c}{Conv + Spectral} \\
\cmidrule{2-3} \cmidrule{5-6} 
& $F-score$ & accuracy  && $F-score$ & accuracy \\ \midrule
Nodule\-level\\
\hspace{3mm}Ben/Mal & 64.3 (6.6)   & 67.0 (7.3)  && 66.0 (6.8) & 68.9 (7.5)  \\
\hspace{3mm}Ben/L/Met  & 43.3 (7.1) & 55.2 (8.4) && 44.4 (7.5) & 56.0 (9.7)  \\
Scan\-level\\
\hspace{3mm}Ben/Mal & 77.8 (10.8) & 79.4 (9.8) && 81.7 (12.3) & 82.9 (10.7) \\
\hspace{3mm}Ben/L/Met & 67.8 (10.8) & 75.0 (7.7)  && 70.0 (10.3) & 78.0 (9.1)  \\
\bottomrule
\end{tabular}
\label{table:ClassificationResults}
\end{table}

\begin{table}[!ht]\centering
\caption{Confusion matrix for nodule-level conventional and spectral data.}
\begin{tabular}{cr*{3}{H}} \toprule
 \multicolumn{2}{c}{}& \multicolumn{3}{c}{Predicted} \\
\cmidrule{3-5} 
&  & \multicolumn{1}{c}{Benign}& 
\multicolumn{1}{c}{Primary Lung } &
\multicolumn{1}{c}{Metastases}  \\ 
\midrule
\multirow{3}{*}{\rotatebox{90}{True}} & Benign & 480 & 54  & 192  \\
& Primary Lung & 88 & 48  & 121   \\
& Metastases& 224 & 104 & 468   \\
\bottomrule
\end{tabular}
\label{table:ConfMatrixSpectNodule}
\end{table}
\begin{table}[!ht]\centering
\caption{Confusion matrix for scan-level conventional and spectral data.}
\begin{tabular}{cr*{3}{E}} \toprule 
 \multicolumn{2}{c}{}& \multicolumn{3}{c}{Predicted} \\
\cmidrule{3-5} 
&  & \multicolumn{1}{c}{Benign}& 
\multicolumn{1}{c}{Primary Lung } &
\multicolumn{1}{c}{Metastases}  \\ 
\midrule
\multirow{3}{*}{\rotatebox{90}{True}} & Benign & 109 & 7  & 8  \\
& Primary Lung & 9 & 18  & 5   \\
& Metastases& 10 & 6 & 32   \\
\bottomrule
\end{tabular}
\label{table:ConfMatrixSpectScan}
\end{table}

Table~\ref{table:ConfMatrixSpectNodule} and~\ref{table:ConfMatrixSpectScan} show the confusion matrices for the nodule- and scan-level classifications respectively obtained from the combination of conventional and spectral features. The results in these tables are a summation of the classification results of each fold during cross-validation. Because the confusion matrices for only the conventional features differ only slightly, these are listed in Appendix~\ref{appendix:results}.

Ideally we would like to have an independent validation of our methodology. Therefore we plan to apply our methodology on the dataset used in~\cite{Nishio2018}. These results will be made available in a next version of this paper.

\section{Discussion} \label{discussion}
In this study, we proposed a methodology to detect lung nodules and classify them based on the origin of the primary tumor. The aim of developing both detection and classification is that ultimately these can be combined in an end-to-end pipeline. To classify based on tumor origin we used a pre-trained malignancy predictor, trained on the LIDC-IDRI, as a feature extractor for our own dataset. Furthermore, we studied the effect of adding spectral features on the classifier performance.

Our detection network was trained on a small (3.2 cm) and large scale (6.4 cm). For both testing and validation subset the combination of both scales performed better than each scale individually, confirming our hypothesis that two scales outperform a single scale. Part of this effect can be attributed to the fact that a combination of two identical models, each stochastically trained, usually performs better than individual predictions~\citep{ju2018relative}. However, we think that specifically the use of two different scales also improves classification because the small scale incorporates more detail whereas the large scale ensures that sufficient background information is included. 

We compared our results obtained from evaluation on the LIDC-IDRI to the top performing teams of the LUNA2016 challenge~\citep{LUNA2016}. For high FP rates our network outperformed the LUNA2016 teams, but for low FP rates we obtained significantly lower performance.

The detection method is developed to be integrated in an end-to-end pipeline for detection and classification. Therefore, we prefer slightly higher detection FP rates over very low rates, as it is important that all significant nodules are included in the classification stage. We propose using a threshold resulting in FP rates around radiologists performance, around 1 FP/scan~\cite{armato2009assessment}, but we recommend that further research is required to establish a definite suitable benchmark.

At 1 FP/scan, our proposed detection method performs equally well to the top contestants of the LUNA2016 challenge, showing that our system obtained state-of-the art performance. It should be noted that we did not perform cross-validation for the evaluation of our detection network, which is performed in the LUNA2016 evaluation, possibly causing minor differences in the evaluation of the results.  

To investigate the generalization ability of our network we applied it to the spectral dataset. On this dataset the detection algorithm performed very poorly, suggesting that the proposed network might not be able to generalize well. 
However, there are some key differences between the two datasets which could contribute to the poor performance. Firstly, the spectral dataset contained more scans of lungs with severe abnormalities which were less present in the LIDC-IDRI data. Secondly, our dataset has been annotated by one radiologist compared to four in the LIDC-IDRI database. As the sensitivity of a single radiologist is only somewhere between 50\% and 80\%~\citep{armato2009assessment}, it is likely that some nodules were missed during annotation resulting in a higher number of FPs. Furthermore, there exists disagreement between radiologists on whether something is a nodule, resulting in even more differences between annotations in the LIDC-IDRI and spectral data. Finally, for evaluation on the LIDC-IDRI database a list of irrelevant findings was used, reducing the number of FPs significantly.

Although the LIDC-IDRI provides a useful framework for the development and evaluation of new networks, the poor performance on the spectral dataset suggests that training on more diverse datasets might be necessary to apply these kind of detection models clinically. As we observed in our dataset, clinical data usually contains more abnormalities which are often excluded in research databases. This makes it challenging to develop nodule detection methods suitable to be used in the clinic.

We used a network with a similar architecture as for detection for the malignancy regression of nodules in the LIDC-IDRI dataset. We were able to classify 91.22\% of the nodules within 1 score point, obtaining a performance comparable to~\citep{Hussein2017}. In their work they used a multi task learning approach with six other nodule attributes to predict the nodule malignancy. Our work suggests that the same performance can be achieved with only using the malignancy score. Among observers the mean error from the mean for the malignancy score is 0.55. At much higher accuracies it will be difficult to to evaluate new systems correctly as no actual ground truth is available, resulting in arguable evaluation.

We used the regression network as feature extractor on the spectral dataset, classifying the features with SVM on nodule- and scan-level. We observed that the element-wise maximum performed best in aggregating nodules into scan-level features. The disadvantage of this method is that it lacks interpretability as it is not completely clear how individual nodule features are combined into one vector. Because neural activations are higher for more important features, the element-wise maximum could be considered as the combination of the most pronounced features from each nodule.

The spectral features in this study are the extracted features from a virtual high and low mono-energetic representation, which were added to the conventional feature vectors using concatenation. We classified the spectral and conventional features vectors as a two- and three-class problem: benign/malignant and benign/primary lung/metastases. The combination of conventional and spectral features achieved slightly higher classification scores in both experiments. However, using a paired-t test on the folds of the cross-validation, these differences showed to be not significant. A possible reason for this is that we only added the mono-energetic views as spectral features. These views show only slight differences with the conventional views, which might be not enough to classify on. Using the Compton scattering and photoelectric effect component might be more promising, as these actually incorporate a physical process. However, our feature extractor was not able to improve classification using these reconstructions (Appendix~\ref{appendix:results}), most likely because these have a quite different appearance then the conventional images. Although not significant, the obtained metrics from the spectral data show the best results.

Using both conventional and spectral features we obtained a classification accuracy of 78\% for the scan-level predictions, and 56\% for the nodule-level predictions. This shows clearly that scan-level classification performed better than nodule-level. Especially the predictions for primary lung nodules are poor for nodule-level (Table~\ref{table:ConfMatrixSpectNodule}), as less than a quarter of the actual lung nodules is predicted to be a lung nodule. The superior scan-level predictions could be explained by the fact that the most prominent attributes are combined into one prediction, yielding more accurate predictions. Furthermore, individual nodules might not always contain specific characteristics resulting in mis-classifications of these nodules. 

To the best of our knowledge, only one other study classifies nodules as benign, primary lung cancer and metastases~\citep{Nishio2018}. In this study a dataset of 1240 patients is used, of which for each patient the most representative nodule is selected. This can thus be considered as a combination of scan- and nodule-level classification. By fine-tuning a pre-trained 2D CNN they obtained an accuracy of 68\%. Because evaluation was not performed on the same dataset, the performance can not be compared directly, but gives an indication of our performance. To make a more direct comparison, we are in touch with the authors of this papers to evaluate our code on their dataset. However, these results are not yet available.

Achieving higher scan-level performance than a fine-tuned network as in~\citep{Nishio2018}, we showed that a pre-trained feature extractor on malignancy can be used for classification based on primary tumor origin. This demonstrates that corresponding features are important for malignancy prediction and for primary tumor classification. The main advantage of using an off-the-shelf feature extractor is that less data is necessary, as only a SVM is fitted on the data. Especially for medical datasets, where usually only small datasets are available, this is an important consideration. Furthermore, it eliminates the need for time-consuming fine-tuning of a network and is easily applicable to new datasets.

A limitation of this study is that the regression network used for malignancy prediction was only trained on a scale of 3.2 cm which was large enough to completely include all nodules in the LIDC-IDRI ($<$ 3 cm). However, the spectral dataset contained nodules larger than 3.2 cm, actually defined as masses, which were only partly visible to the network. In future work, a larger scale could be included to classify these masses. Another shortcoming of this study is that it does not yet apply detection and classification in an end-to-end pipeline. It would be interesting to observe classifier performance using the detected locations from the network as input. Ultimately, this would eliminate the necessity of nodule annotations overcoming the problems associated with different annotators per dataset.

Our pre-trained network succeeds in extracting the necessary features for classification based on primary tumor origin, but yet still lacks the ability to exploit the spectral features to the fullest. We suggest that future research looks into training a multi-stream network, in which the inputs of the different streams are spectral representations. Examples of the application of multi-stream networks are in computer vision, where they are used to combine temporal and spatial information for action recognition~\citep{simonyan2014two} and in medical image analysis, where they can be used to fuse different MRI modalities~\citep{nie2016fully}. Using a multi-stream architecture, the network might be able to learn how to combine the spectral information better. Most likely a larger dataset is necessary to train such a classification network. In this study we were not able to classify the metastases separately as melanoma and colorectal cancer. We think that by making full use of the spectral features, in future work it should be possible to make this differentiation.

\section{Conclusion} \label{conclusion}
In conclusion, we have shown that a similar network can be used for both detection and malignancy regression on the LIDC-IDRI, achieving state-of-the art performance. The trained regression neural network was transferred to be used as a feature extractor on a new dataset. We showed that using these features a classification based on primary tumor origin (benign/primary lung/metastases) can be made, obtaining an accuracy of 78\% for scan-level predictions. The addition of spectral views did result in slightly higher classification scores, however these differences were not statistically significant.

\section*{Acknowledgment}

We gratefully acknowledge the support of NVIDIA Corporation with the donation of the Titan Xp GPU used for this research. We would also like to thank the Diagnostic Image Analysis Group group in Nijmegen for providing us with their annotation software.

\renewcommand*{\bibfont}{\footnotesize}
\bibliographystyle{IEEEtranN}
\bibliography{articles.bib}

\setcounter{figure}{0}  
\setcounter{table}{0}
\clearpage

\onecolumn
\appendices
\section{Excluded Scans}
\label{appendix:exclusion}

\renewcommand{\arraystretch}{1.2}
\begin{table}[ht]
\caption{Excluded scans from the spectral dataset. The comments were made during nodule annotation by our radiologist.}
\begin{minipage}[t]{0.5\textwidth}\centering
	\vspace{0pt}
	\begin{tabular*}{0.8\linewidth}{@{\extracolsep{\fill}}rr@{}} \toprule
	Scan Number& Reason for Exclusion \\
  	\midrule 
	0 & Nodules not found \\
	3 & No colorectal metastases found \\
	11 & No real nodules \\
	15 & Too much uncertainty \\
	20 & Nodules unclear \\
	21 & Nodules unclear \\
	38 & CT abdomen \\
	39 & - \\
	42 & More consolidations than nodules \\
    54 & Missing slices  \\
	60 & Not annotated (skipped) \\
	63 & Not suitable \\
	65 & CT abdomen \\
	68 & Not suitable \\
	70	&  Not suitable \\
    79	& CT head \\
	80	& Diagnosis unsure \\
	81 &	Not suitable \\
	82 &	Not suitable \\
	84 &	Not suitable \\
	85	& CT abdomen \\
	96 & Not able to segment in atelactasis \\
	97	& Not able to segment in atelactasis \\
	105	& CT feet \\
	125	& Not able to segment nodule\\
	127	& - \\
	128	& Not able to segment nodule\\
	129	& Not able to segment nodule \\
	139	& - \\
	143	& - \\
	145	& Not clear what tumor is and what not \\
	147	& Same as other scan (146) \\
	149	& - \\

	\bottomrule
	\end{tabular*}
\end{minipage} 
\hfill
\begin{minipage}[t]{0.5\textwidth}\centering
	\vspace{0pt}
	\begin{tabular*}{0.8\linewidth}{@{\extracolsep{\fill}}rr@{}}
    \toprule
	Scan Number& Reason for Exclusion \\
  \midrule 
  	152	& Deviates to much to interpret \\
    155 & Missing slices \\
	158	& - \\
  	160	& No real lung nodules \\
	161	& - \\
	162	& - \\
	174	& Mucus plugs \\
	178	& - \\
	180	& - \\
	193	& - \\
	195	& - \\
	204	& - \\
	205	& - \\
	209	& Nodules unclear \\
	210	&- \\
	225	& No measurable metastases \\
	228	& Only half a scan \\
	230	& Scan too irregular \\
	239	& - \\
	246	& - \\
	248	& Not able to segment nodule\\
	250	& CT neck \\
	252	& - \\
	253	& - \\
	255	& Diagnosis unsure \\
	256	& - \\
	257	& - \\
	258	& - \\
	262	& - \\
	266	& - \\
	274	& - \\
	279	& Infection \\
	& \\
	\bottomrule
	\end{tabular*}
\end{minipage}
\end{table}

\clearpage

\section{Adapted and missing segmentations}
\label{appendix:segmentations}

\begin{table}[h]\centering
\caption{ Segmentations from the spectral dataset which were incorrect using the standard segmentation method, if no solution is listed no segmentation is applied to these scans.}
\begin{tabular}{@{}lrr@{}} \toprule
Scan Number & Reason for Error & Solution  \\ \midrule
12 & Tracheotomy    & - \\
40 & Tracheotomy   & slice[:34] = 0     \\ 
62 & Holes in segmentation (deviating lung structure) & -\\
88  & Head on CT & - \\
89 & Head on CT  & slice[:35, 319:] = 0 \\
91 & Holes in segmentation (deviating lung structure) & - \\
100 & Tracheotomy & slice[:35, 319:] = 0 \\
102	& Holes in segmentation (deviating lung structure) & - \\
104 & Head on CT & - \\
116 & Head on CT & slice[:22, 316:] = 0 \\
133 & Head on CT & slice[:186, 444:] = 0 \\
146 & Head on  CT & slice[:57, 342:] = 0 \\
156 & Head on CT & slice[:38] = 0 \\
157 & Holes in segmentation (deviating lung structure) & - \\
185 & Head on CT	& - \\
188 & Head on CT & slice[:215, 464:] = 0 \\
203 & Holes in segmentation (deviating lung structure) & - \\
223 & Holes in segmentation (deviating lung structure) & dilation kernel = 15 mm \\
248 & Head on CT & - \\
258 & Holes in segmentation (deviating lung structure) & -\\
262 & Holes in segmentation (deviating lung structure) & - \\
263 & Head on CT & slices[:172, 419:] = 0 \\
274 & Holes in segmentation (deviating lung structure) & - \\
276 & Holes in segmentation (large mass attached to wall) & - \\

\bottomrule
\end{tabular}
\label{table:MissingSegms}
\end{table}

\clearpage
\section{Additional Results}
\begin{figure*}[!ht]
	\centering
	\setlength\figureheight{9 cm}
	\setlength\figurewidth{0.6\textwidth}
\begin{tikzpicture}
\def\colorModel{RGB}
\definecolor{color4}{RGB}{255,127,0}
\definecolor{color3}{RGB}{152,78,163}
\definecolor{color2}{RGB}{77,175,74}
\definecolor{color1}{RGB}{55,126,184}
\definecolor{color0}{RGB}{227,26,28}
\begin{axis}[
height=\figureheight,
legend cell align={left},
legend entries={{Znet},{Aidence},{JianPeiCAD},{MOT},{Our Method}},
legend style={at={(0.97,0.03)}, anchor=south east, draw=white!80.0!black},
tick align=outside,
tick pos=left,
width=\figurewidth,
xmode=log,
x grid style={white!69.01960784313725!black},
xlabel={average FP/scan},
xmin=0.125, xmax=16,
xtick={0.125,0.25, 0.5,1 ,2,4,16},
xticklabels={0.125,0.25, 0.5,1 ,2,4,16},
y grid style={white!69.01960784313725!black},
ylabel={sensitivity},
ymin=0.4, ymax=1,
ytick={0.0,0.2,0.4,0.6,0.8,1},
yticklabels={0.0,0.2,0.4,0.6,0.8,1.0}
]
\addlegendimage{ line width=1.5 pt,color4}
\addlegendimage{ line width=1.5 pt,  color3}
\addlegendimage{line width=1.5 pt,  color2}
\addlegendimage{line width=1.5 pt, color1}
\addlegendimage{line width=1.5 pt, color0}
\addplot [ thick, color4]
table [row sep=\\]{%
0.062039417	0.588994528 \\
0.066796357	0.59470615\\
0.077853505	0.611551945\\
0.090477476	0.634616864\\
0.105750594	0.650867631\\
0.123842912	0.662490249\\
0.143671457	0.680059945\\
0.167465828	0.693209907\\
0.19520805	0.704431607\\
0.243962006	0.721586738\\
0.283579819	0.733580225\\
0.331458873	0.748368726\\
0.385254495	0.754081625\\
0.450346932	0.77391877\\
0.563230091	0.792441601\\
0.679831772	0.805860301\\
0.786314543	0.818279576\\
0.916624695	0.826608884\\
1.068481552	0.837348519\\
1.245500256	0.847927466\\
1.451937269	0.855185519\\
1.692599009	0.862175757\\
1.973157338	0.868994594\\
2.300274793	0.87454935\\
2.681688663	0.878807885\\
3.126247604	0.884726867\\
3.644636366	0.888717588\\
4.248772805	0.895332887\\
4.953136333	0.901037618\\
5.774509077	0.904546274\\
6.703723095	0.909967844\\
7.967119594	0.915429369\\
9.539908807	0.921921056\\
11.12193436	0.925269024\\
12.96587809	0.930384564\\
14.80301055	0.933428158\\
15.71945618	0.93397238\\
16.0 0.93397238 \\
};
\addplot [thick, color3]
table [row sep=\\]{%
0.062882689	0.474159662\\
0.064675625	0.485638064\\
0.068810665	0.504287656\\
0.075749964	0.517189736\\
0.084337131	0.531525271\\
0.094415938	0.554472297\\
0.100452427	0.573121889\\
0.108712963	0.588897725\\
0.122422927	0.600360483\\
0.137835716	0.621871119\\
0.152573302	0.64338469\\
0.170806752	0.666331716\\
0.19230575	0.689277764\\
0.209302258	0.705052622\\
0.225313456	0.702169088\\
0.255176732	0.712195456\\
0.285695161	0.730836248\\
0.307483566	0.739436005\\
0.354217373	0.748024028\\
0.405715632	0.760919263\\
0.469933324	0.780989599\\
0.514485121	0.785280189\\
0.609612472	0.802475791\\
0.718208558	0.822543194\\
0.926469188	0.841159541\\
1.110493214	0.848305308\\
1.208805481	0.85690311\\
1.346057013	0.862626177\\
1.568185989	0.875518479\\
1.80658039	0.882671091\\
2.240363333	0.88837558\\
2.874301414	0.895509613\\
3.368199522	0.899788469\\
3.969177375	0.90693717\\
5.607364348	0.912619169\\
7.921890203	0.916865758\\
10.10743127	0.916823712\\
12.39445884	0.916788511\\
14.52540976	0.916761133\\
15.72499724	0.915312032\\
16.0 0.915312032\\
};
\addplot [ thick, color2]
table [row sep=\\]{%
0.062105046	0.533013484 \\
0.065929724	0.547305472\\
0.076825224	0.576684965\\
0.089762098	0.586982305\\
0.105242419	0.61037511\\
0.116669188	0.620255697\\
0.124075934	0.633325071\\
0.141767407	0.647603291\\
0.166163316	0.663323304\\
0.1791168	0.678511425\\
0.203625258	0.685784525\\
0.237338885	0.701184124\\
0.277026056	0.715555077\\
0.323831901	0.724896241\\
0.375894426	0.736991738\\
0.43813458	0.751855709\\
0.51068865	0.765862675\\
0.595284002	0.777512878\\
0.694018201	0.779575338\\
0.809090867	0.784101687\\
0.943136453	0.794627071\\
1.099475593	0.801028119\\
1.281559166	0.814499457\\
1.494055936	0.818811555\\
1.741728888	0.824891226\\
2.030496107	0.830006766\\
2.367203693	0.83367611\\
2.748072542	0.840126087\\
3.217401075	0.840693422\\
3.729865584	0.846966882\\
4.34841382	0.850122023\\
5.126690815	0.850800636\\
5.943654751	0.853603226\\
6.929326054	0.856790505\\
8.078187353	0.861745357\\
9.417755033	0.865414701\\
10.98002338	0.866352342\\
12.80156584	0.866807918\\
14.82004491	0.872535278\\
15.82721633	0.872248713\\
};
\addplot [ thick, color1]
table [row sep=\\]{%
0.062508755	0.489950165 \\
0.069759714	0.512908517\\
0.081306908	0.530075689\\
0.098805206	0.553502485\\
0.118385483	0.575964356\\
0.152689162	0.603193171\\
0.200303451	0.631854464\\
0.244334464	0.640814023\\
0.284774941	0.658677511\\
0.331944428	0.670863342\\
0.386962586	0.678174958\\
0.451026266	0.694110185\\
0.525749208	0.704689132\\
0.615065645	0.70905419\\
0.832910223	0.725552721\\
0.970845099	0.739184748\\
1.131765917	0.746121424\\
1.31936924	0.75268316\\
1.538094885	0.758387891\\
1.79935348	0.763909013\\
2.09025418	0.771939865\\
2.458151913	0.776397848\\
2.871811066	0.782175024\\
3.861251957	0.787206433\\
4.52826297	0.78540244\\
5.265593403	0.787997543\\
6.155161782	0.787755324\\
7.111722983	0.788308254\\
8.317663873	0.792644518\\
9.572433705	0.796258935\\
10.96680738	0.796235467\\
13.18102612	0.797867767\\
14.73784827	0.797591401\\
15.75999361	0.7976083\\
16.0 0.7976083 \\
};

\addplot [thick, color0]
table [row sep=\\]{%
152.155172413793	0.990697674418605 \\
109.155172413793	0.976744186046512 \\
84.2011494252874	0.976744186046512 \\
67.3275862068966	0.976744186046512 \\
59.8735632183908	0.972093023255814 \\
53.5114942528736	0.972093023255814 \\
16.5057471264368	0.967441860465116 \\
11.0977011494253	0.958139534883721 \\
6.47701149425287	0.958139534883721 \\
2.81609195402299	0.93953488372093 \\
1.5632183908046	0.897674418604651 \\
1.36206896551724	0.888372093023256 \\
1.18390804597701	0.869767441860465 \\
0.994252873563218	0.837209302325581 \\
0.890804597701149	0.823255813953488 \\
0.758620689655172	0.8 \\
0.591954022988506	0.758139534883721 \\
0.310344827586207	0.637209302325581 \\
0.218390804597701	0.53953488372093 \\
0.103448275862069	0.32093023255814 \\
0.0689655172413793	0.246511627906977 \\
0.0172413793103448	0.144186046511628 \\
0	0 \\
};    
\end{axis}
\end{tikzpicture}
    \label{fig:FROCLUNA}
	\caption{FROC curves on the LIDC-IDRI database of our method compared to the 4 top contestants of the LUNA2016 challenge. The FROC curve of our method is the performance on the testing subset for a combination of small and large scale.  The x-axis is shown in logarithmic scale to emphasize the differences at low FP rates. The curves from the LUNA2016 contestants were extracted from the figure in \citep{LUNA2016}.  }
\end{figure*}
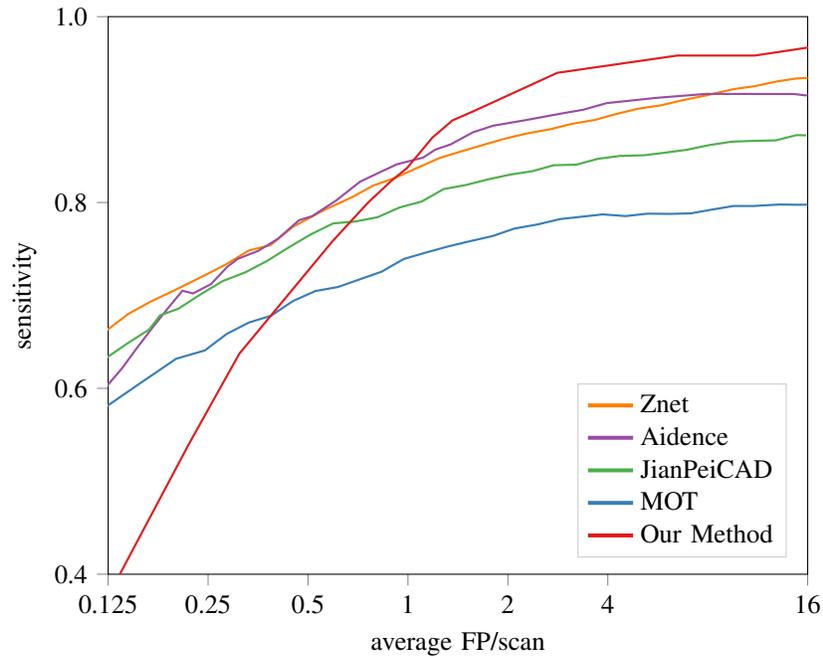
\label{appendix:results}
\begin{figure*}[!ht]
	\centering
	\setlength\figureheight{9 cm}
	\setlength\figurewidth{0.6\textwidth}
\begin{tikzpicture}

\definecolor{color0}{RGB}{31,120,180}
\definecolor{color1}{RGB}{51,160,44}
\definecolor{color2}{RGB}{227,26,28}
\begin{axis}[
height=\figureheight,
legend cell align={left},
legend entries={{3.2 cm},{6.4 cm},{Both Scales}},
legend style={at={(0.97,0.03)}, anchor=south east, draw=white!80.0!black},
tick align=outside,
tick pos=left,
width=\figurewidth,
x grid style={white!69.01960784313725!black},
xlabel={average FP/scan},
xmin=0, xmax=25,
xtick={0, 5,10,15,20,25},
xticklabels={0, 5,10,15,20,25},
y grid style={white!69.01960784313725!black},
ylabel={sensitivity},
ymin=0, ymax=1,
ytick={0,0.2,0.4,0.6,0.8,1},
yticklabels={0.0,0.2,0.4,0.6,0.8,1.0}
]
\addlegendimage{ line width=1.5 pt, mark=*,mark size=0.7,color0}
\addlegendimage{ line width=1.5 pt, mark=*,mark size=0.7, color1}
\addlegendimage{line width=1.5 pt, mark=*,mark size=0.7, color2}

\addplot [thick, color0, mark=*, mark size=0.7, mark options={solid}]
table [row sep=\\]{%
43.7429906542056	0.911691542288557 \\
39.5794392523364	0.901741293532338 \\
35.6869158878505	0.893034825870647 \\
31.0233644859813	0.887437810945274 \\
27.9906542056075	0.878109452736318 \\
21.8831775700935	0.851990049751244 \\
18.3598130841122	0.838308457711443 \\
16.5467289719626	0.825248756218905 \\
14.6635514018692	0.80410447761194 \\
13.3504672897196	0.788557213930348 \\
12.3177570093458	0.775497512437811 \\
10.9439252336449	0.759328358208955 \\
9.03738317757009	0.711442786069652 \\
8.21028037383178	0.687810945273632 \\
7.24299065420561	0.663557213930348 \\
6.71028037383178	0.641791044776119 \\
6.14485981308411	0.618781094527363 \\
5.63551401869159	0.591417910447761 \\
5.38785046728972	0.572761194029851 \\
3.31775700934579	0.373134328358209 \\
0	0 \\
};

\addplot [thick, color1, mark=*, mark size=0.7, mark options={solid}]
table [row sep=\\]{%
40	0.876244 \\
23.4906542056075	0.876243781094527 \\
19.0794392523364	0.850746268656716 \\
16.7616822429907	0.834577114427861 \\
15.1028037383178	0.817164179104478 \\
13.8878504672897	0.80410447761194 \\
11.3364485981308	0.770522388059702 \\
9.66355140186916	0.754353233830846 \\
8.81308411214953	0.73818407960199 \\
7.72897196261682	0.70771144278607 \\
6.76635514018692	0.677238805970149 \\
5.96261682242991	0.649875621890547 \\
4.80841121495327	0.608208955223881 \\
3.00467289719626	0.480721393034826 \\
2.32242990654206	0.388059701492537 \\
1.41588785046729	0.256840796019901 \\
0.962616822429907	0.192786069651741 \\
0.775700934579439	0.130597014925373 \\
0.47196261682243	0.0802238805970149 \\
0.397196261682243	0.0541044776119403 \\
0	0 \\
};

\addplot [thick, color2, mark=*, mark size=0.7, mark options={solid}]
table [row sep=\\]{%
42.411214953271	0.924751243781095 \\
34.4532710280374	0.912313432835821 \\
28.9018691588785	0.89365671641791 \\
22.2336448598131	0.875621890547264 \\
13.7476635514019	0.838930348258706 \\
8.89252336448598	0.776741293532338 \\
7.62616822429907	0.753731343283582 \\
6.98130841121495	0.738805970149254 \\
6.19158878504673	0.712686567164179 \\
5.49532710280374	0.691542288557214 \\
5.05140186915888	0.659825870646766 \\
4.31308411214953	0.621268656716418 \\
3.07476635514019	0.51865671641791 \\
2.41121495327103	0.444651741293532 \\
1.74766355140187	0.330223880597015 \\
1.27570093457944	0.236318407960199 \\
0.97196261682243	0.173507462686567 \\
0.724299065420561	0.0988805970149254 \\
0.551401869158878	0.0652985074626866 \\
0	0 \\
};





\end{axis}

\end{tikzpicture}
    \label{fig:FROCspect}
	\caption{FROC curves of the network performance on the spectral dataset for all scales. It should be noted that the y-axis has a different range than the figures in the main paper.}
\end{figure*}
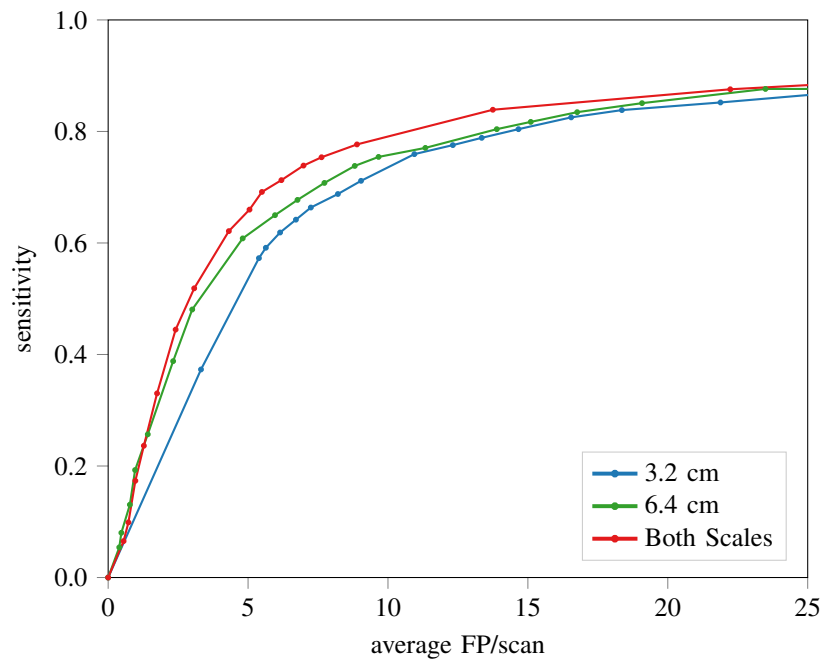
\begin{figure*}[!ht]
	\centering
	\setlength\figureheight{7.5 cm}
	\setlength\figurewidth{0.48\textwidth}
	\input{AccuracyBoxplots.tikz}
    \label{fig:FROCspect2}
	\caption{Box plots of the accuracy obtained by 10-fold cross-validation. Results are shown both for scan-level and nodule-level predictions and for both the conventional features and a combination of conventional and spectral features.}
\end{figure*}
\begin{figure*}[!ht]
	\centering
	\setlength\figureheight{7.5 cm}
	\setlength\figurewidth{0.48\textwidth}
	\input{CS_PE_F1macro.tikz}
    \label{fig:FROCspect2}
	\caption{Box plots of the F-score obtained by 10-fold cross-validation. Results are shown both for scan-level and nodule-level predictions and for both the conventional features and a combination of conventional and Compton scattering (CS) / photoelectric effect (PE) features. }
\end{figure*}
\clearpage
\begin{figure*}[!t]
	\centering
	\setlength\figureheight{7.5 cm}
	\setlength\figurewidth{0.48\textwidth}
	\input{accuracy_CSPE.tikz}
    	\label{fig:CSPE_acc}
	\caption{Box plots of the accuracy obtained by 10-fold cross-validation. Results are shown both for scan-level and nodule-level predictions and for both the conventional features and a combination of conventional and Compton scattering (CS) / photoelectric effect (PE) features.}
\end{figure*}

\begin{table}\centering
\caption{Confusion matrix for nodule-level conventional data.}
\begin{tabular}{@{}cr*{3}{H}} \toprule
 \multicolumn{2}{c}{}& \multicolumn{3}{c}{Predicted} \\
\cmidrule{3-5} 
&  & \multicolumn{1}{c}{Benign}& 
\multicolumn{1}{c}{Primary Lung } &
\multicolumn{1}{c}{Metastases}  \\ 
\midrule
\multirow{3}{*}{\rotatebox{90}{True}} & Benign & 477 & 64  & 185  \\
& Lung & 93 & 45  & 119   \\
& Metastases& 230 & 105  & 461   \\
\bottomrule
\end{tabular}
\label{table:ConfusionNodule}
\end{table}
\begin{table}\centering
\caption{Confusion matrix for scan-level conventional data.}
\begin{tabular}{cr*{3}{E}} \toprule
 \multicolumn{2}{c}{}& \multicolumn{3}{c}{Predicted} \\
\cmidrule{3-5} 
&  & \multicolumn{1}{c}{Benign}& 
\multicolumn{1}{c}{Primary Lung } &
\multicolumn{1}{c}{Metastases}  \\ 
\midrule
\multirow{3}{*}{\rotatebox{90}{True}} & Benign & 102 & 11  & 11  \\
& Lung & 6 & 20  & 6   \\
& Metastases& 10 & 7  & 31   \\
\bottomrule
\end{tabular}
\label{table:ConfusionScan}
\end{table}
\begin{table}[!ht]\centering
\caption{Confusion matrix for nodule-level conventional data with all classes.}
\begin{tabular}{cr*{4}{H}} \toprule
 \multicolumn{2}{c}{}& \multicolumn{4}{c}{Predicted} \\
\cmidrule{3-6} 
&  & \multicolumn{1}{c}{Benign} & 
\multicolumn{1}{c}{Primary Lung } &
\multicolumn{1}{c}{Colorectal}  &
\multicolumn{1}{c}{Melanoma} \\ 
\midrule
\multirow{4}{*}{\rotatebox{90}{True}} & Benign & 518 & 74  & 73 & 61   \\
& Lung & 108 & 53  & 64 & 32 \\
& Colorectal & 168 & 75 & 89 & 153  \\
& Melanoma & 114 & 27 & 122 & 48  \\
\bottomrule
\end{tabular}
\label{table:onfusionAllDataNodules}
\end{table}
\begin{table}[!ht]\centering
\caption{Confusion matrix for scan-level conventional data with all classes.}
\begin{tabular}{cr*{4}{E}} \toprule
 \multicolumn{2}{c}{}& \multicolumn{4}{c}{Predicted} \\
\cmidrule{3-6} 
&  & \multicolumn{1}{c}{Benign}& 
\multicolumn{1}{c}{Primary Lung } &
\multicolumn{1}{c}{Colorectal}  &
\multicolumn{1}{c}{Melanoma} \\ 
\midrule
\multirow{4}{*}{\rotatebox{90}{True}} & Benign & 99 & 13  & 5 & 7   \\
& Lung & 7 & 18  & 4 & 3 \\
& Colorectal & 10 & 2  & 3 &  13\\
& Melanoma & 3 & 4 & 8 & 5 \\
\bottomrule
\end{tabular}
\label{table:ConfusionAllDataScan}
\end{table}
\vfill

\end{document}